\documentclass[10pt,twocolumn]{article}

\usepackage{framed,multirow}

\usepackage{amssymb}
\usepackage{latexsym}

\usepackage{url}
\usepackage{xcolor}
\usepackage{wrapfig}
\usepackage{placeins}
\definecolor{newcolor}{rgb}{.8,.349,.1}

\usepackage{graphicx}
\usepackage{amsmath}

\usepackage{color}

\usepackage{siunitx}
\usepackage{textcomp}
\usepackage{gensymb}
\usepackage{subcaption}
\usepackage{hyperref}
\usepackage{algorithmic, algorithm}
\usepackage{adjustbox}

\makeatletter
\@namedef{ver@everyshi.sty}{}
\makeatother

\usepackage{ctable}

\usepackage[
backend=biber,
style=alphabetic,
]{biblatex}
\newcommand{\ie}{i.e. }

\newcommand{\rebuttalAdd}[1]{#1}
\newcommand{\rebuttalCor}[2]{#2}

\date{}

\newcommand\blfootnote[1]{%
  \begingroup
  \renewcommand\thefootnote{}\footnote{#1}%
  \addtocounter{footnote}{-1}%
  \endgroup
}

\begin{document}

\thispagestyle{empty}

\title{Physically-admissible polarimetric data augmentation for road-scene analysis}

\author{Cyprien Ruffino$^{1,*}$, Rachel Blin$^{1,*}$, Samia Ainouz$^1$, Gilles Gasso$^1$,\\ Romain H\'erault$^1$, Fabrice Meriaudeau$^2$, St\'ephane Canu$^1$
\vspace{.3cm}\\
1- Normandie Univ, UNIROUEN, UNIHAVRE, INSA Rouen, LITIS\\ 76~000 Rouen, France
\vspace{.1cm}\\
2- University of Burgundy, UBFC, ImViA, 71200 Le Creusot, France
}

\maketitle

\begin{abstract}
Polarimetric imaging, along with deep learning, has shown improved performances on different tasks including scene analysis. However, its robustness may be questioned because of the small size of the training datasets. Though the issue could be solved by data augmentation,
polarization modalities are subject to physical feasibility constraints unaddressed by classical data augmentation techniques. To address this issue, we propose to use CycleGAN, an image translation technique based on deep generative models that solely relies on unpaired data, to transfer large labeled road scene datasets to the polarimetric domain. We design several auxiliary loss terms that, alongside the CycleGAN losses, deal with the physical constraints of polarimetric images. The efficiency of this solution is demonstrated on road scene object detection tasks where generated realistic polarimetric images allow to improve performances on cars and pedestrian detection up to 9\%. The resulting constrained CycleGAN is publicly released, allowing anyone to generate their own polarimetric images.\blfootnote{Corresponding author: cyprien.ruffino@insa-rouen.fr \\ $*$ Equal contribution}

\end{abstract}

%\linenumbers

%% main text

\section{Introduction}

Road scene analysis is a vital task for driving assistance systems. Robust road users detection including vehicles, pedestrians, animals, is a precondition for a secure navigation. The complexity and variability of outdoor illumination conditions make RGB-based detection algorithms poor and limited, especially when they are faced to strong reflections, noises or bad weather conditions. In contrast, polarization encoded images are a non-conventional modality where each reflected wave light from a pixel is strongly linked to the physical properties of the surface. The significant interest resides in the fact that polarimetric imaging is a rich modality that enables to characterize an object by its reflective properties. In a polarimetric image, each pixel encodes information regarding the object's roughness, its orientation and its reflection \cite{wolff1995polarization}. Applications of polarimetric imaging range from indoor autonomous navigation \cite{berger2017depth}, depth map estimation \cite{Zhu_2019_CVPR}, 3D objects reconstruction \cite{morel2006active} to differentiation of healthy and unhealthy cervical tissues in order to detect cancer at an early stage \cite{rehbinder2016ex}. Also, recently, polarization imaging was exploited in autonomous driving applications either to enhance car detection \cite{fan2018polarization}, road mapping and perception \cite{aycock2017polarization} or to detect road objects in adverse weather conditions \cite{blin2019road}. The key element of this work comes from the combination of polarization richness and deep learning-based detection. However, these applications are still characterized by the reduced size of the available training databases which restrains them from using deep neural networks. To overcome this limitation, one possible solution is the generation of polarimetric encoded images. 

Generative Adversarial Networks (GAN) \cite{goodfellow2014generative,wang2019generative} are powerful deep generative models used to implicitly learn complex data distributions and to generate realistic samples from them. In its standard form, a GAN consists of two models: a generator which maps samples drawn from a latent low-dimensional distribution (usually uniform or Gaussian distributions) to high-dimensional points expected to follow the sought data distribution, and a discrimination model which discriminates the real samples from the generated ones \cite{goodfellow2014generative}. GAN have proven remarkable for various application domains including image generation \cite{arjovsky17a}, image-to-image translation \cite{isola2017image, zhu2017unpaired, hoffman18a_cycada} or image attribute manipulation \cite{Antipov_face_aging} to name a few. 

Arguably most of the impressive achievements of GAN were obtained for color images. A body of work attempted to extend GAN architectures to other uncommon imaging domains. For instance, some existing methods rely on CycleGAN \cite{zhu2017unpaired}, an image-to-image translation network, to generate infrared road scenes from RGB counterpart images \cite{zhang2018synthetic}, to produce thermal images for person re-identification \cite{Kniaz_2018_ECCV_Workshops} or for infrared image colorization \cite{mehri2019colorizing}. In the same vein, data augmentation in the field of medical imaging was achieved \cite{nie2017medical} by transforming MRI inputs into pseudo-CT images. Following the previous stream of work, this paper contributes to generative models for non-conventional imaging techniques. Specifically we propose a generative model framework to produce realistic polarimetric images. 

Unlike RGB, LiDAR or infrared image generation which mostly responded to visual qualitative constraints, unless some learnable knowledge constraints are enforced (see \cite{hu2018deep} for pose conditional person image generation), sampling polarization images is more challenging. Indeed, this imaging technique comes with physical admissibility constraints on the pixels of an image \cite{bass1995handbook}. To be physically feasible, each pixel entry of such an image should satisfy some physical constraints related to light polarization principle and to the calibration setup of the acquisition devices.
Therefore, we propose a set of constraints that, when satisfied, ensure that the generated images are physically feasible. Based on these constraints, we formulate our problem of polarimetric image generation as a domain-transfer learning problem under physical constraints to ensure that the generated images are valid. 
We propose a learning framework based on CycleGAN \cite{zhu2017unpaired}, which enables unpaired image-to-image translation with relatively few images, to which we add constraints for handling the physical polarization constraints during training. This allows for circumventing the expensive labelling step by transferring a source labelled RGB dataset to the polarimetric domain while keeping the shapes and contents of the source image unchanged. 
 
We demonstrate the effectiveness of our constrained-output CycleGAN on the KITTI dataset \cite{geiger2012we} as well as on the Berkeley Deep Drive dataset (BDD100K) \cite{xu2017end}, which are commonly used for object detection in road scenes. Using the generated polarization-encoded images to train a deep object detector, we witness an improvement of the detection performances of cars and pedestrians which are of great interest for autonomous driving applications. 

To summarize, the contributions of this paper are:
\begin{itemize}
    \item as far as our knowledge goes, we propose the first framework for generating physically admissible polarization-encoded images starting from RGB images,
    \item we propose an extension of CycleGAN which allows to generate polarimetric-encoded images while handling the physical constraints that should be satisfied by the pixels of the generated images,
    \item we show that, when plugged into the training procedure of an object detector for pre-training, the generated images help improving the detection performances.
\end{itemize}

The remainder of the paper is organized as follows: 
the polarization formalism and the physical constraints it involves are first presented. Then, the CycleGAN approach is described and a way to take into account these physical constraints during the training process of the CycleGAN for generating polarimetric images is investigated. Experimental evaluations are conducted; they aim to translate RGB images from KITTI and BDD100K datasets into polarimetric images. Finally, the generated images are exploited to boost the performances of an object detection network. The code for the experiments and the trained models are available at: \url{https://anonymous.4open.science/r/4a83820e-9c65-417c-af3a-ab2979d6e2e8/}

%%%%%%%%%%%%%%%%%%%%%%%%%%%%%%%%%%%%%%%%%%%%%%%%%%%%%%%%%%%%%%%%%%%%%%%%%%%%%%%%%%%%%%%%%%%%%%%%%%%

\section{Framework}

This section introduces the polarization formalism, the GAN approach and the CycleGAN principles. It focuses on the formulation of the proposed modelling framework, namely the learning of a CycleGAN with output constraints. A solution approach and the related learning principle are presented.

\rebuttalAdd{\subsection{Physical principles of polarimetry}
\label{physical_prop}

Polarization is the direction of propagation of the electrical field of the light wave. The electrical field of the wave propagating in $z$ direction at time $t$ is characterized by the scalar components of the vector
%given by its scalar components components $\textbf{E}= (E_x(z,t), E_y(z, t))$, where 
\begin{equation}
    \textbf{E} = \begin{bmatrix}
        E_x(z, t) \\
        E_y(z, t) \\
        E_z(t)
    \end{bmatrix} = \begin{bmatrix}
        E_{0x}\cos(\omega t - kz) \\
        E_{0y}\cos(\omega t - kz + \phi) \\
        0
    \end{bmatrix}, 
\end{equation}
with $E_{0x}$ and $E_{0y}$ the maximum amplitudes of each component. $\phi$ is the phase shift between the two components $E_x(z,t)$ and $E_y(z, t)$, $\omega$ is the angular frequency, $k$ represents the wave number, directly related to the wavelength $\lambda$. Figure \ref{fig:propagation} depicts the different parameters.

\begin{figure}
    \centering
    \includegraphics[width=\linewidth]{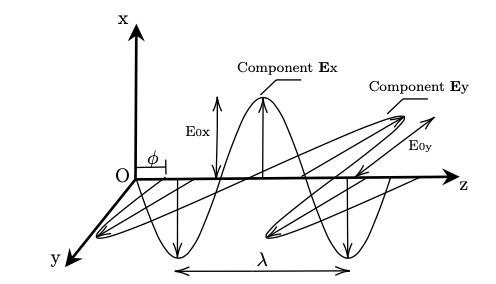}
    \caption{Propagation principle of the electromagnetic wave, with $E_{0x}$ and $E_{0y}$ the amplitudes, $\lambda$ the wavelength and $\phi$ the phase shift of the wave}
    \label{fig:propagation}
\end{figure}

The two components of the electrical field could be combined by eliminating the temporal term $\omega t$ in the following way : 

\begin{equation}
    \left(\frac{E_x(t)}{E_{0x}}\right)^2 + \left(\frac{E_y(t)}{E_{0y}}\right)^2 - 2\frac{E_x(t)E_y(t)}{E_{0x}E_{0y}}\cos(\phi) = \sin^2(\phi)
    \label{eq:propagation}
\end{equation}

If $\phi$ is constant over the time, Equation \ref{eq:propagation} defines an ellipse meaning that the electrical field of a propagating wave has an elliptical behaviour as illustrated in Figure \ref{fig:ellipse}.

\begin{figure}
    \centering
    \includegraphics[width=\linewidth]{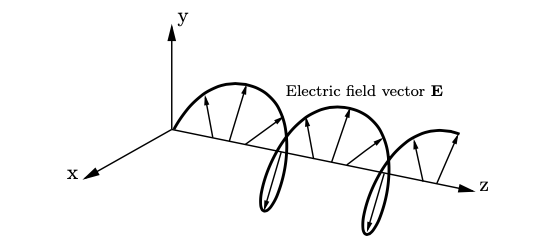}
    \caption{The elliptical behaviour of the electrical field of an electromagnetic light wave in the 3D space: the polarization ellipse}
    \label{fig:ellipse}
\end{figure}

According to $\sin(\phi)$, three states of polarization could be observed. If $\sin(\phi)>0$ the polarization is said to be right-handed, otherwise, if $\sin(\phi)<0$, the polarization is left handed. If $\sin(\phi)=0$, the trajectory of the wave is linear and the polarization is known as linear polarization. If $\sin(\phi)= \pm 1$, the polarization of the wave is respectively right and left circular. For these three cases the wave is totally polarized, otherwise, if $\phi$ is not constant over the time, the wave is unpolarized \cite{bass1995handbook}.  Usually, in an uncontrolled environment, a light wave is a combination of a totally polarized part and an unpolarized one, this is the reason why the natural light is known to be partially polarized. 

\subsection {Mathematical representation}

There are different ways to describe the polarization of the light and its interaction with the media. The Stokes parameters are one of them to describe the polarization state of an electromagnetic wave, especially in the general case of a partially polarized wave form. The Stokes vector $[S_0, S_1, S_2, S_3]$ contains four polarization parameters where each component brings us information on the reflected light wave as follows:
\begin{itemize}
    \item $S_0$ represents the total intensity of the light which is not a polarization property. $S_{0}>0$ is always positive, which means that every natural object, when illuminated, reflects a light,
    \item $S_1$ defines the amount of the polarized light in the wave horizontally or vertically polarized, meaning that the electrical field represented by figure \ref{fig:ellipse} moves in a horizontal or a vertical way,
    \item $S_2$ is the amount of the polarized light oriented at $45^\circ$ and $-45^\circ$ and,
    \item $S_3$ represents the amount of the circular polarization. 
\end{itemize}
In the case of outdoor scenes, the sunlight is unpolarized and it is physically proved that when an unpolarized light wave is being reflected, it becomes partially linearly polarized. Its polarization depends on the normal surface and the refractive index of the material it impinges \cite{wolff1995polarization}. The circular component is thus neglected and set to zero in all applications concerning non controlled environments \cite{morel2006active}. We focused in our work on the only linearly polarized light represented by the three first components of the Stokes vector, namely $[S_0, S_1, S_2]$.

In order to access to observable optical field, the two components $E_x(t)$ and $E_y(t)$ are averaged over the time. The polarization ellipse of Equation \ref{eq:propagation} average in case of linear polarization $(sin(\phi)=0)$ is given by the following equation:
\begin{equation}
    \frac{\left< E_x(t)^2 \right>}{E^2_{0x}} + \frac{\left< E_y(t)^2 \right>}{E^2_{0y}} -2\frac{\left< E_x(t)E_y(t) \right>}{E^2_{0x}E^2_{0y}} = 0.
    \label{eq:linear_polar}
\end{equation}
The averaged values of the electrical components are:
\begin{align}
    \left< E^2_x(t)\right> &= \frac{1}{2}E^2_{0x} \\
    \left< E^2_y(t)\right> &= \frac{1}{2}E^2_{0y} \\
    \left< E_x(t)E_y(t) \right> &= \frac{1}{2}E_{0x}E_{0y}.
\end{align}
By replacing the time averaged terms in Equation \ref{eq:linear_polar}, the elliptical equation can be rewritten in the following form

\begin{equation}
    (E^2_{0x} +E^2_{0y}) - (E^2_{0x} -E^2_{0y}) - (2E_{0x}E_{0y}) = 0.
    \label{eq:rearranged}
\end{equation}
The linear Stokes vector is finally derived from Equation \ref{eq:rearranged} as 

\begin{equation}
    S = \begin{bmatrix}
    S_0\\S_1\\S_2
    \end{bmatrix} = 
    \begin{bmatrix} 
        E^2_{0x} + E^2_{0y} \\
        E^2_{0x} - E^2_{0y} \\
        2E_{0x}E_{0y}.
\end{bmatrix}
\end{equation}

One salient physical property, obtained from the Stokes parameters, is the degree of linear polarization (DOP) \cite{ainouz2013adaptive} defined by:
\begin{equation}
DOP = \frac{\sqrt{S_1^2+S_2^2}}{S_0} \enspace.
\end{equation}
The $DOP \in [0,1]$ (or equivalently between 0 and 100\% ). It refers to the rate of the polarized part in the light wave. It is equal to 1 for a totally polarized light, 0 for unpolarized light and between 0 and 1 for partially polarized light. 

Equation \ref{eq:rearranged} describes the totally polarized wave. By construction of the Stokes vector, the total energy $S_0^2$ in this case is equal to the partial energy $S_1^2 + S_2^2$ meaning that the light is totally composed of polarized waves linearly oriented. In case of partially polarized wave, the wave contains an unpolarized part, the equality is transformed to an inequality meaning that $ S_0^2\geqslant S_1^2 + S_2^2$. 
From this property, the Stokes vector is said to be physically admissible if and only if the two following conditions are met: 
 \begin{equation}
    S_0 > 0
\quad \mbox{ and } \quad 
    S_0^2 \geqslant S_1^2 + S_2^2 \enspace.
    \label{eqn:stokes_constraint_S0}
\end{equation}
This condition means also that the DOP never exceeds 1, otherwise the partial energy of the wave may exceed its total energy, which is, physically, an impossible phenomenon. 
\subsection {Stokes imaging}

Polarimetric images are obtained by computing the Stokes vector related to each pixel. The acquisition principle is based on a device composed of a polarizer oriented at an angle $\alpha$ between the object and the sensor \cite{Wang_2019_CVPR}. At least three acquisitions with three different angles are required to get the Stokes parameters. The reflected light from the object, represented by the unknown Stokes vector, passes through the polarizer before reaching the camera. 
In our work, we use a Polarcam 4D Technology polarimetric camera, providing simultaneously four images respectively obtained with four different linear polarizers oriented at $(\alpha_i)_{i=1:4} =$ (0\degree, 45\degree, 90\degree, 135\degree). The polarimetric camera measures an intensity $I_{\alpha_i}$ of the scene for each angle $\alpha_i$. The relationship between the Stokes vector $S$ and the intensities $I(\alpha_i)_{i=1:4}$ reaching the camera is given by: 
\begin{equation}
    I_{\alpha_i} = \tfrac{1}{2}\begin{bmatrix}
    1 & \cos(2\alpha_i) & \sin(2\alpha_i)
    \end{bmatrix} \begin{bmatrix}S_0 \\ S_1 \\ S_2\end{bmatrix} \enspace,
    \forall i=1, 4 %\nonumber 
\end{equation}
\noindent that is:
  \begin{equation}
    I = AS \enspace, 
    \label{eqn:IAS}
  \end{equation}
\noindent where $I = \begin{bmatrix} I_0 & I_{45} & I_{90} & I_{135}\end{bmatrix}^\top$ refers to the four intensities according to each angle of the polarizer $(\alpha_i)_{i=1:4}$ and $A \in \mathbb{R}^{4\times 3}$, to the calibration matrix of the polarization camera, defined as: 
\begin{equation}
A = \frac{1}{2} {\begin{bmatrix}
1 & \cos(2\alpha_1) & \sin(2\alpha_1) \\
1 & \cos(2\alpha_2) & \sin(2\alpha_2) \\
1 & \cos(2\alpha_3) & \sin(2\alpha_3) \\
1 & \cos(2\alpha_4) & \sin(2\alpha_4)
\end{bmatrix}}
\\
 = \frac{1}{2} {\begin{bmatrix}
1 & 1 & 0 \\
1 & 0 & 1 \\
1 & -1 & 0 \\
1 & 0 & -1
\end{bmatrix}} \enspace. %\nonumber
\end{equation}
An example of the different intensities for the same scene is shown\footnote{\rebuttalAdd{Since the polarimetric images can be very dark, we manually adjust their brightness in the Figures \ref{fig:polar_overview intensities}, \ref{fig:experimental_setup},  \ref{fig:overview_polarCycle}, \ref{fig:polar_example} and \ref{fig:reco_polar}}}  in Figure~ \ref{fig:polar_overview intensities}. 

\begin{figure}[!th]
\centering
\begin{subfigure}{0.24\columnwidth}
  \centering
  \includegraphics[width=\linewidth]{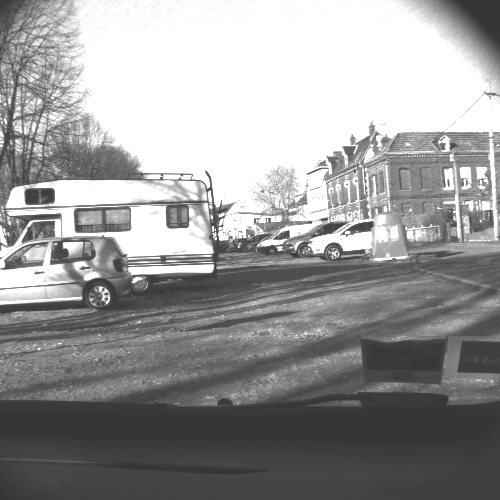}
\end{subfigure}%
\begin{subfigure}{0.24\columnwidth}
  \centering
  \includegraphics[width=\linewidth]{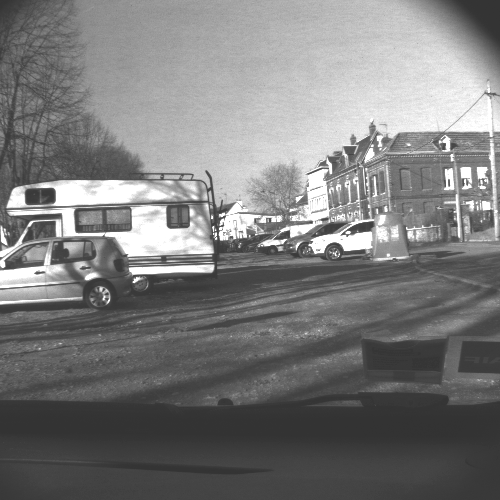}
\end{subfigure}%
\begin{subfigure}{0.24\columnwidth}
  \centering
  \includegraphics[width=\linewidth]{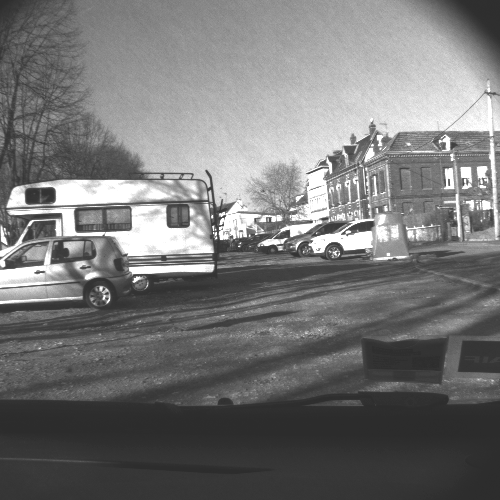}
\end{subfigure}%
\begin{subfigure}{0.24\columnwidth}
  \centering
  \includegraphics[width=\linewidth]{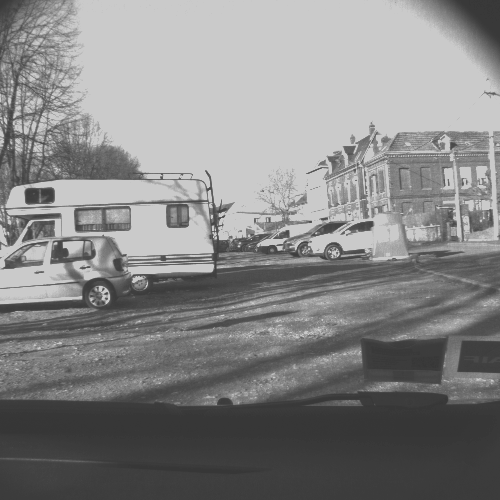}
\end{subfigure}
\caption{Example of a polarimetric image. From left to right, the intensities corresponding to the polarizer rotation angles 0$\degree$, 45$\degree$, 90$\degree$ and 135$\degree$.}
\label{fig:polar_overview intensities}
\end{figure}
To get the unknown Stokes parameters from the measured intensities (see Equation (\ref{eqn:IAS})), we require $\tilde{A} = (A^\top A)^{-1} A^\top \in \mathbb{R}^{3\times 4}$ the pseudoinverse of the matrix $A$. The relationship between $S$ and $I$ is then defined by:
\begin{eqnarray}
S & = & \tilde{A}I = 
\begin{bmatrix}
1 & 0 & 1 & 0 \\
1 & 0 & -1 & 0 \\
0 & 1 & 0 & -1
\end{bmatrix}
\begin{bmatrix} 
I_0 \\
I_{45} \\
I_{90} \\
I_{135}
\end{bmatrix} 
 = 
\begin{bmatrix} 
I_0 + I_{90} \\
I_0 - I_{90} \\
I_{45} - I_{135} 
\end{bmatrix}
\label{eqn:stokes2} \enspace.
\end{eqnarray}
Combining Equations (\ref{eqn:IAS}) and (\ref{eqn:stokes2}), we have: 
\begin{equation}
    I = A\tilde{A}I \enspace,
\end{equation}
\noindent which is satisfied if and only if:
\begin{equation}
I_0 + I_{90} = I_{45} + I_{135} \enspace.
\label{eqn:physics}
\end{equation}

Stokes images should then satisfy two main conditions: the physical admissibility constraints in \eqref{eqn:stokes_constraint_S0}
and the calibration constraint given by \eqref{eqn:physics}. The generated polarimetric images have to comply with these essential constraints.}

\subsection{Unpaired image-to-image translation with CycleGAN}

Given two domains $X$ and $Y$, unpaired image-to-image translation is the task of learning the mapping functions $M_{XY} : X \rightarrow Y$ and $M_{YX} : Y \rightarrow X$ using unpaired samples $x_i \in X$ with $i \in [1..N]$ and $y_j \in Y$ with $j \in [1..M]$. 
An effective approach to achieve the task is CycleGAN \cite{zhu2017unpaired}. 
It consists in learning the two mapping models $M_{XY}$ and $M_{YX}$ by combining the objective function of the standard Generative Adversarial Network (GAN) \cite{goodfellow2014generative} with a Cycle-Consistency loss function. The adversarial cost related to the GAN serves for training the models to generate samples that will match the target domain distribution, while the Cycle-Consistency cost ensures that the learned models are able to correctly reconstruct an original image (of the source domain) from a generated one.

Formally a GAN is composed of a generative model $G : Z \rightarrow X$ which maps a known distribution $p_Z$, usually normal or uniform, to the unknown distribution $p_X$ of the samples and a discrimination model $D : X \rightarrow [0,1]$. The generator $G$ attempts to fool the discriminator $D$, which in turn tries to distinguish a real sample from a sample generated by the model $G$. Learning a GAN amounts to solve the following problem:
\begin{equation}
     G^*, D^* = \arg\min_G\max_D L_{GAN}(D, G) \enspace,
\end{equation}
\noindent with:
\begin{align}
     L_{GAN}(D, G) =& \mathop{\mathbb{E}}_{x \sim p_X} \Big[\log (D(x))\Big] \\
     &+\mathop{\mathbb{E}}_{z\sim p_Z} \Big[\log (1-D(G(z)))\Big] \enspace, 
\end{align}.

CycleGAN trains the two models $M_{XY}$ and $M_{YX}$ by using solely unpaired real samples $x$ $\in X$ and $y \in Y$ respectively drawn according to the (unknown) distributions $p_X$ and $p_Y$ as input. It also learns two discrimination networks $D_X: X \rightarrow [0,1]$ and $D_Y: Y\rightarrow [0,1]$ able to detect generated samples from real ones in the domains $X$ and $Y$ respectively. CycleGAN relies on the Least-Squares variant of GAN \cite{mao2017least} and considers the following adversarial costs:
\begin{align}
    L_{GAN}(D_Y,& M_{XY}) = \nonumber \\
    &\mathop{\mathbb{E}}_{y \small{\sim} p_Y} \Big[(D_Y(y) - 1)^2\Big] + \mathop{\mathbb{E}}_{x\small{\sim}p_X}\Big[D_Y(M_{XY}(x))^2\Big] \enspace, \\
    L_{GAN}(D_X,& M_{YX}) = \nonumber \\
    &\mathop{\mathbb{E}}_{x \small{\sim}p_X} \Big[(D_X(x) - 1)^2\Big] + \mathop{\mathbb{E}}_{y\small{\sim}p_Y} \Big[D_X(M_{YX}(y))^2\Big] \enspace.
\end{align}
In order to ensure the cyclic consistency, \ie both the compositions $M_{XY}~\circ~M_{YX}$ and $M_{YX}~\circ~M_{XY}$ are identity functions, a $\ell_1$ reconstruction error term is devised for the mapping models: 
\begin{align}
&L_{reco}(M_{XY}, M_{YX}) = \nonumber \\
&\mathop{\mathbb{E}}_{y \small{\sim} p_Y}||y - M_{XY}(M_{YX}(y))||_1 +\mathop{\mathbb{E}}_{x \small{\sim} p_X}||x - M_{YX}(M_{XY}(x))||_1
\enspace.
\end{align}
Gathering all these elements leads to the objective function: 
\begin{align}
    L_{CycleGAN}(D_X, &D_Y, M_{XY}, M_{YX}) = L_{GAN}(D_Y, M_{XY}) + \nonumber \\ 
     &L_{GAN}(D_X, M_{YX})+\lambda L_{reco}(M_{XY}, M_{YX})
    \enspace, \label{eqn:Lcyclegan}
\end{align}

\noindent where $\lambda > 0$ is an hyper-parameter that controls the reconstruction term. Training a CycleGAN consists in solving, via alternate gradient descent, the following minmax problem: 

\begin{align}
    M_{XY}^*,& M_{YX}^*, D_X^*, D_Y^* = \nonumber \\
    &\arg\min_{\substack{M_{XY}\\M_{YX}}}\max_{\substack{D_X\\D_Y}} L_{CycleGAN}(D_X, D_Y, M_{XY}, M_{YX}) \enspace. \label{eq:cycleGAN}
\end{align}

\subsection{Proposed approach}
\label{polarGAN}

As discussed above, our main goal is to learn a generative model able to produce realistic polarization-based images starting from RGB images. As our proposed solution to this problem, we adopt the image-to-image translation framework and extend it to account for the constraints a polarimetric image must fulfill. 

To generate a polarimetric image from an RGB one, we propose to use the CycleGAN approach to learn the translation models $M_{XY}$ and $M_{YX}$ between $X$ the domain of the polarimetric images and $Y$ the RGB image domain. Let $\hat{I} \in \mathbb{R}^4$ be the intensity vector associated to a pixel of a generated polarimetric image. To be physically admissible, each pixel has to satisfy the admissibility (\ref{eqn:stokes_constraint_S0}) and the calibration (\ref{eqn:physics}) constraints. 
We refer to these polarimetric constraints by $\mathcal{C}_1$, $\mathcal{C}_2$ and $\mathcal{C}_3$ as follows:
\begin{eqnarray}
    \mathcal{C}_1 &:& I = AS\enspace, \nonumber\\
    \mathcal{C}_2 &:& S_0^2 \geqslant S_1^2 + S_2^2 \enspace, \\
    \mathcal{C}_3 &:& S_0 > 0 \enspace. \nonumber
\end{eqnarray}
By design, $S_0$ is always positive as it represents the total intensity reflected from an object. As the last layer of the generation models customary uses the hyperbolic tangent as activation function, each output intensity $\hat I$ is within the range $]-1,1[$ which we scale to $]0,255[$. Hence $\hat{S}_0=\hat{I}_0+\hat{I}_{90}$ (see (\ref{eqn:stokes2})) is ensured to be strictly positive. Therefore, constraint $\mathcal{C}_3$ can be deemed satisfied for the generated polarimetric images. To handle the remaining constraints $\mathcal{C}_1$ and $\mathcal{C}_2$, one could resort to the Lagrangian dual of CycleGAN optimization problem (\ref{eq:cycleGAN}) subject to these constraints. However, this may be computationally expensive, as it requires to optimize four neural networks (respectively the discrimination and the mapping network models) in an inner loop of a dual ascent algorithm. Moreover the overall optimization procedure may not be stable because of the minmax game involved in the CycleGAN learning. 

In order to derive an efficient algorithm to learn CycleGAN under output constraints, we introduce a relaxation of the problem. Instead of strictly enforcing the constraints, we measure how far the generated image pixels are from the feasibility domain through additional cost functions we attempt to minimize.
For the constraint $\mathcal{C}_1$, we propose to use a $\ell_2$ distance between the generated image $M_{YX}$ and $A\hat{S}$ as:
\begin{equation}
L_{\mathcal{C}_1} = \mathop{\mathbb{E}}_{y\sim p_Y} ||M_{YX}(y) - A\hat{S}||_2\enspace,
\label{eqn:ls}
\end{equation}
with $\hat{S}=\begin{bmatrix}
\hat{S_0} & \hat{S_1} & \hat{S_2}
\end{bmatrix}^\top$ the Stokes vector computed from the generated image by $M_{YX}(y)$ using equation \eqref{eqn:stokes2}.
 Similarly, to enforce the constraint $\mathcal{C}_2$, a rectified linear penalty $L_{\mathcal{C}_2}$ is considered, defined as:
\begin{equation}
L_{\mathcal{C}_2} = \mathop{\mathbb{E}}_{y\sim p_Y} \max\left(\hat{S}_1^2 + \hat{S}_2^2 -
\hat{S}_0^2, 0 \right)\enspace.
\label{eqn:lreg}
\end{equation}
The loss $L_{\mathcal{C}_1}$ enforces the respect of the acquisition conditions according to the calibration matrix $A$ while $L_{\mathcal{C}_2}$ pushes the generated images towards respecting the physical admissibility constraint on the Stokes vectors obtained from the generated image.

Gathering all these elements, we train our CycleGAN by optimizing the following objective function:
\begin{equation}
L_{final}= L_{CycleGAN}+\mu L_{\mathcal{C}_1} + \nu L_{\mathcal{C}_2} \enspace.
\label{eqn:lfinal}
\end{equation}
The full training algorithm is summed up in Algorithm \ref{alg:cyclegan_train}
\begin{algorithm}
    \begin{algorithmic}[H]
       \REQUIRE{$X$ and $Y$ two unpaired datasets, $M_{XY}$ and $M_{YX}$ the mapping networks, $D_X$ and $D_Y$ the discrimination models, $m$ the mini-batch size}
        \REPEAT
            \STATE sample a mini-batch $\lbrace x_i \rbrace_{i=1}^m$ from $X$\;
            \STATE sample a mini-batch $\lbrace y_i \rbrace_{i=1}^m$ from $Y$\;
            \STATE update $D_X$ by stochastic gradient descent of
            \STATE \ \ \ \ $ \sum_{i=1}^{m}(D_X(x_i)-1)^2 + (D_X(M_{YX}(y_i)))^2$
            \STATE update $D_Y$ by stochastic gradient descent of
            \STATE \ \ \ \ $ \sum_{i=1}^{m}(D_Y(y_i)-1)^2 + (D_Y(M_{XY}(x_i)))^2$
            \STATE sample a mini-batch $\lbrace x_i \rbrace_{i=1}^m$ from $X$\;
            \STATE sample a mini-batch $\lbrace y_i \rbrace_{i=1}^m$ from $Y$\;
            \STATE update $M_{XY}$ by stochastic gradient descent of
            \STATE \ \ \ \ $ \sum_{i=1}^n (D_Y(M_{XY}(x_i))-1)^2 + \lambda (||x_i - M_{YX}(M_{XY}(x_i))||_1$ \STATE \ \ \ \ \ \ \ \ $+||y_i -M_{XY}(M_{YX}(y_i))||_1)$\;
            \STATE update $M_{YX}$ by stochastic gradient descent of
            \STATE \ \ \ \ $ \sum_{i=1}^n (D_X(M_{YX}(y_i))-1)^2+ \lambda \big[||x_i - M_{YX}(M_{XY}(x_i)||_1 $
            \STATE \ \ \ \ \ \ \ \ $+ ||y_i - M_{XY}(M_{YX}(y_i))||_1\big]$\;
            \STATE \ \ \ \ \ \ \ \ $ + \mu (||M_{YX}(y_i) - A\hat{S}||_2) + \nu \max(\hat{S}_1^2 + \hat{S}_2^2 - \hat{S}_0^2, 0)$
           
        \UNTIL a stopping condition is met
    \end{algorithmic}
    \caption{Constrained CycleGAN training algorithm}
    \label{alg:cyclegan_train}
\end{algorithm}

\begin{figure}[!th] 
    \centering
    \includegraphics[width=\linewidth]{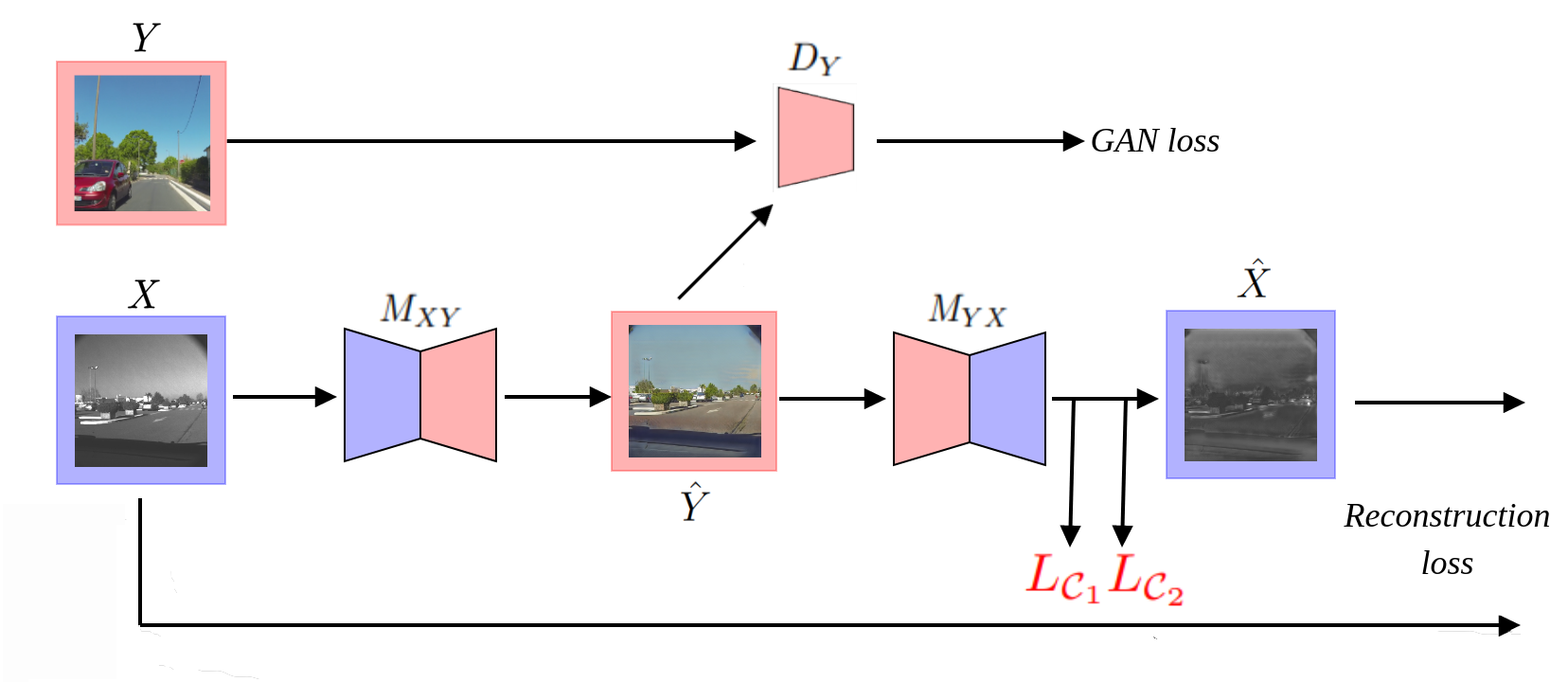}
    \caption{\rebuttalAdd{Training of the $M_{YX}$ model using the CycleGAN training process extended to our constraints. The full training objective consists of    a GAN cost using the discriminator network $D_Y$, a cycle-consistent reconstruction cost which ensures that $M_{XY}(X)$ remains consistent with $X$, and our custom objectives $L_{\mathcal{C}_1}$ and $L_{\mathcal{C}_2}$ (in red).}}
    \label{fig:overview_polarCycle}
\end{figure}

The non-negative hyper-parameters $\mu$ and $\nu \in \mathbb{R}^{+}$ control respectively the balance of calibration and admissibility constraints according to the CycleGAN loss $L_{CycleGAN}$ (see \eqref{eqn:Lcyclegan}). As the values of $L_{\mathcal{C}_1}$ and $L_{\mathcal{C}_2}$ are computed pixel-wisely, we consider their averages over the whole image in the objective function. The training principle of the proposed generative model is illustrated in Figure~\ref{fig:overview_polarCycle}.

\begin{figure*}[!th]
    \centering
    \includegraphics[width=\linewidth]{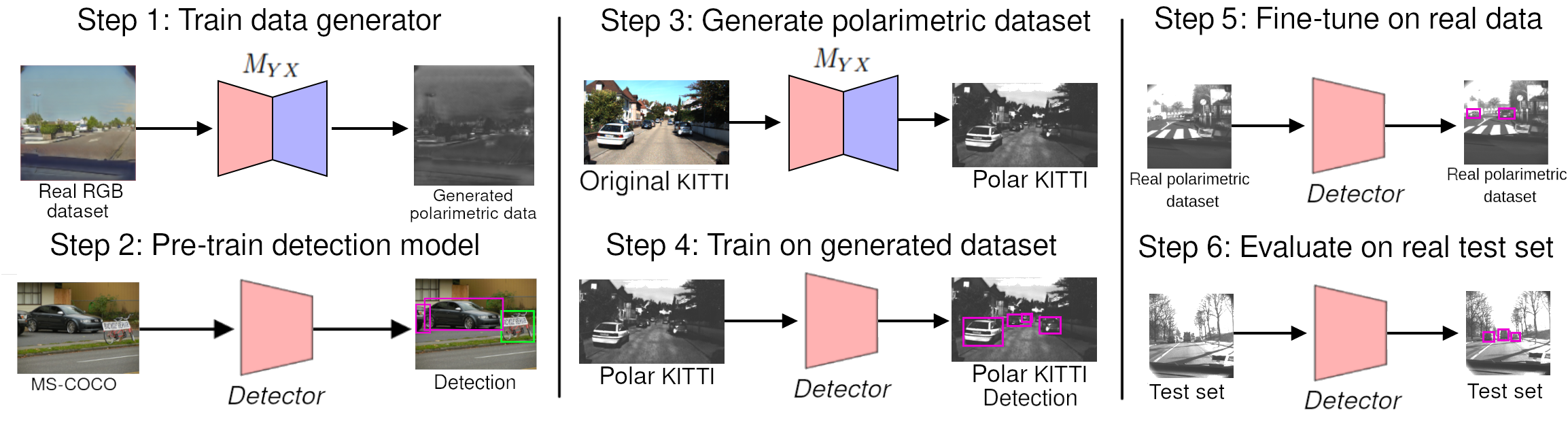}
    \caption{\rebuttalCor{}{Full setup of our approach. For each experiment, we 1): train the generator $M_{YX}$; 2): pre-train the RetinaNet-50 detection network on the MS-COCO dataset; 3): generate a full polarimetric dataset using the trained generator $M_{YX}$; 4): perform a first training on the generated polarimetric KITTI dataset; 5): fine-tune on the real polarimetric dataset; and 6): evaluate the detection model on the test set of the real polarimetric dataset. Note that while the procedure is illustrated with the KITTI dataset, it straightforwardly extends to the BDD100K dataset.}}
    \label{fig:experimental_setup}
\end{figure*}

%%%%%%%%%%%%%%%%%%%%%%%%%%%%%%%%%%%%%%%%%%%%%%%%%%%%%%%%%%%%%%%%%%%%%%%%%%%%%%%%%%%%%%%%%%%%%%%%%%%%%%%%%%%%%%%%%%%%%%%%%%%%%%%%%%

\section{Experimental evaluation}

Hereafter, the experimental setup, including the image generation procedure and its evaluation, is presented. 

\subsection{Polarimetric images generation using CycleGAN} \label{subsec:polar_gen}

\begin{figure}[!th]
\centering
\begin{subfigure}{.2\columnwidth}
  \centering
  \includegraphics[width=\linewidth]{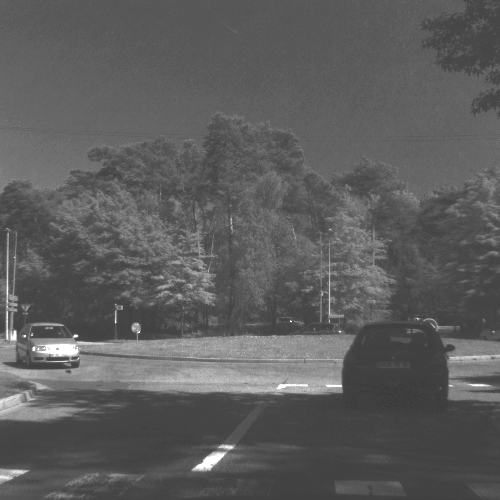}
\end{subfigure}%
\begin{subfigure}{.2\columnwidth}
  \centering
  \includegraphics[width=\linewidth]{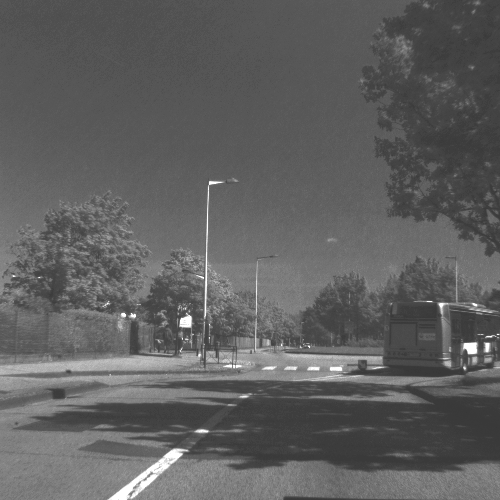}
\end{subfigure}%
\begin{subfigure}{.2\columnwidth}
  \centering
  \includegraphics[width=\linewidth]{images/2474_I0.png}
\end{subfigure}%
\begin{subfigure}{.2\columnwidth}
  \centering
  \includegraphics[width=\linewidth]{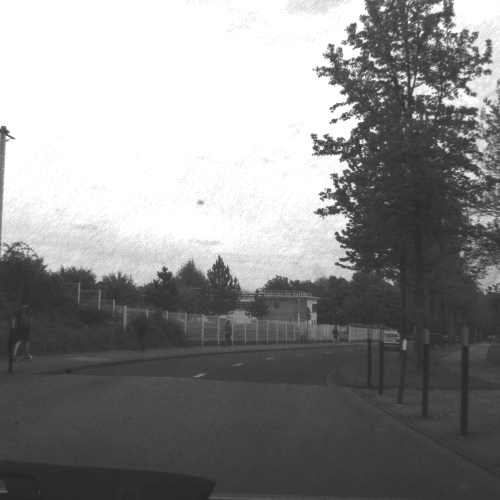}
\end{subfigure}%
\begin{subfigure}{.2\columnwidth}
  \centering
  \includegraphics[width=\linewidth]{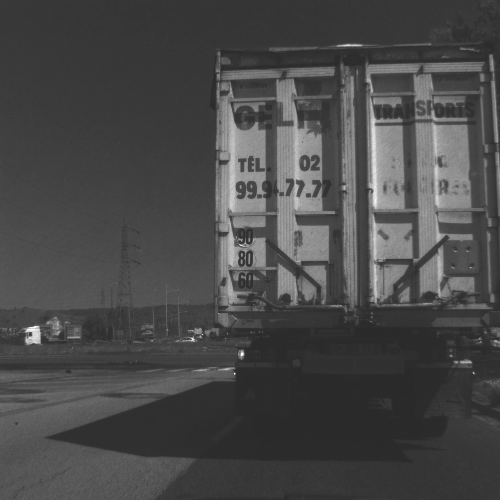}
\end{subfigure}
\begin{subfigure}{.2\columnwidth}
  \centering
  \includegraphics[width=\linewidth]{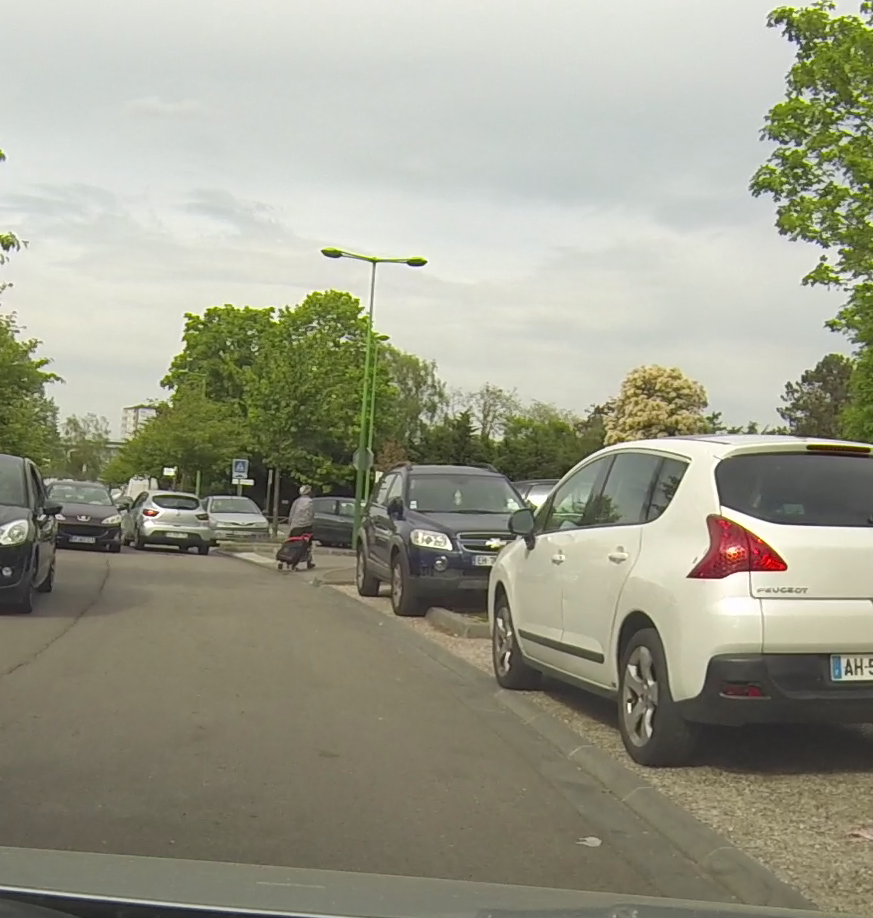}
\end{subfigure}%
\begin{subfigure}{.2\columnwidth}
  \centering
  \includegraphics[width=\linewidth]{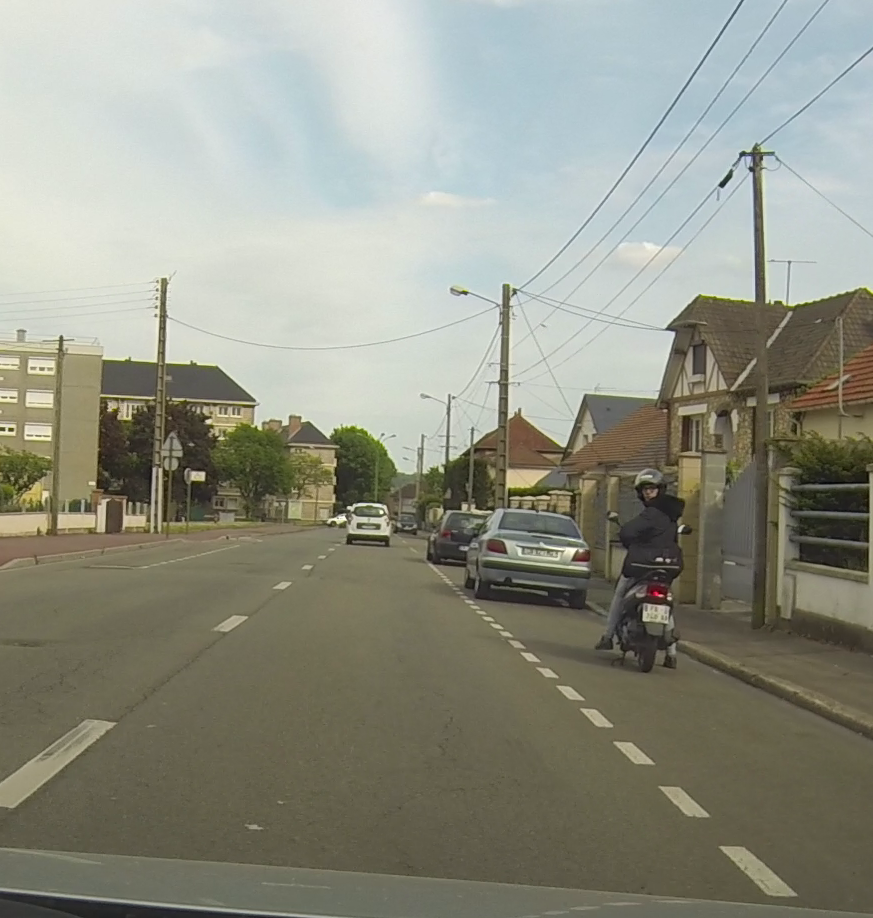}
\end{subfigure}%
\begin{subfigure}{.2\columnwidth}
  \centering
  \includegraphics[width=\linewidth]{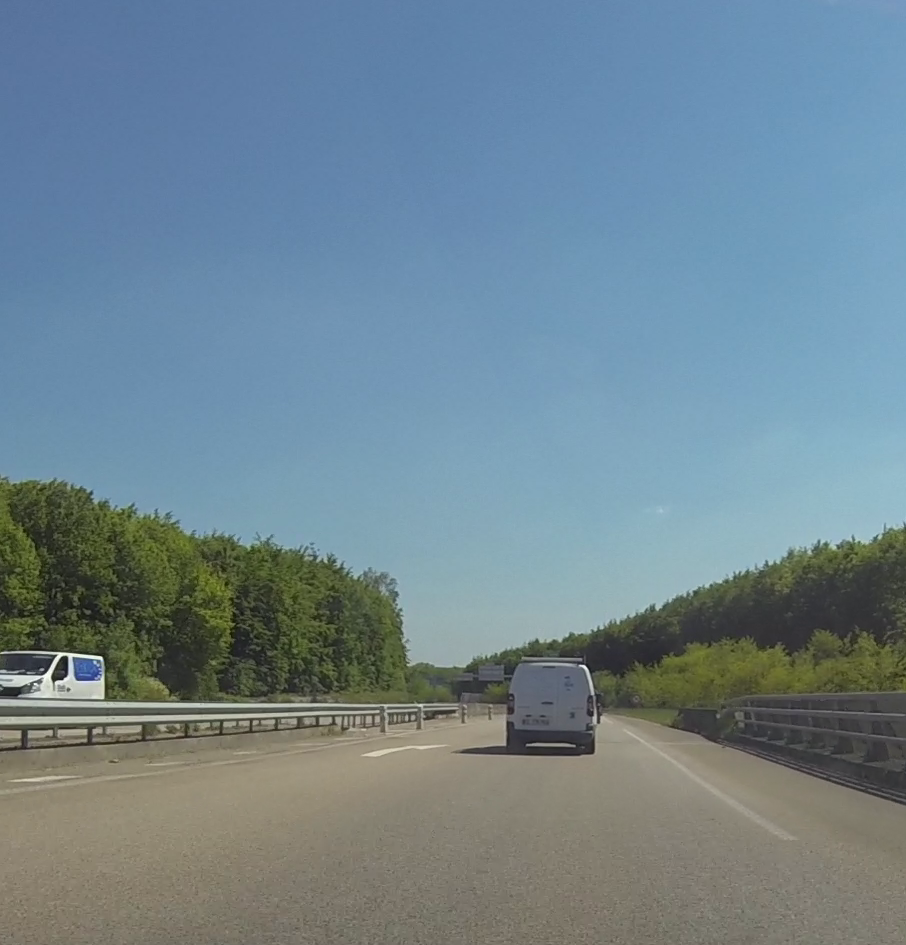}
\end{subfigure}%
\begin{subfigure}{.2\columnwidth}
  \centering
  \includegraphics[width=\linewidth]{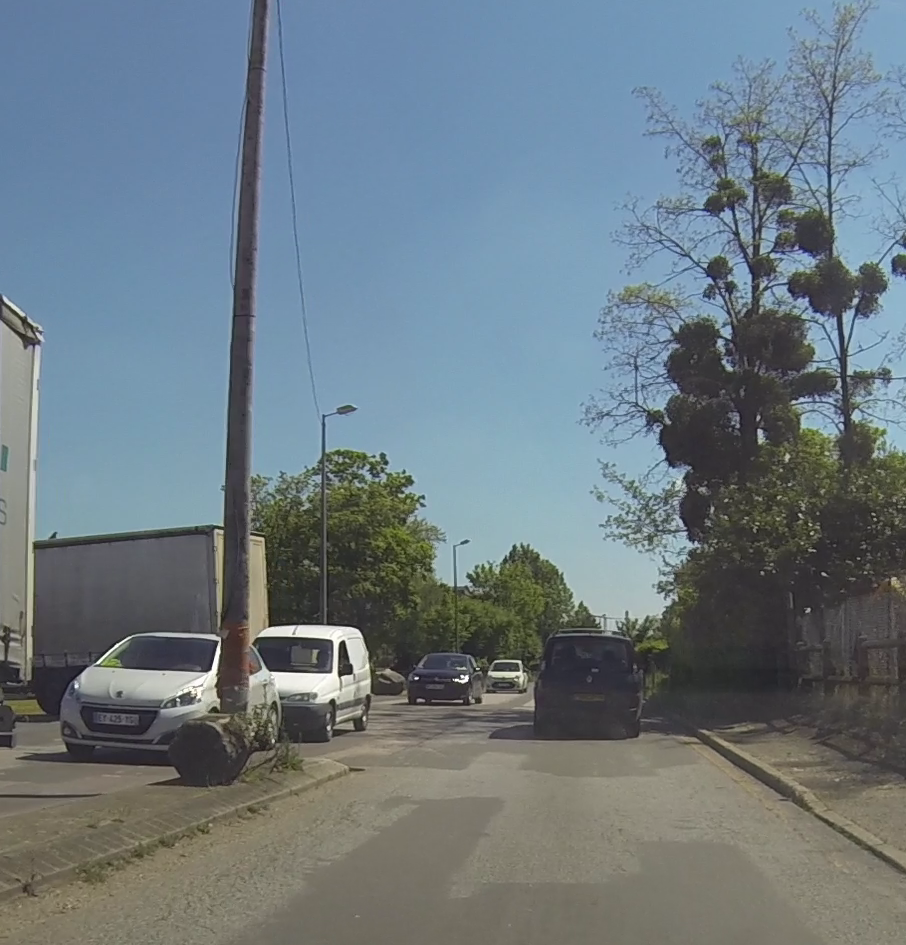}
\end{subfigure}%
\begin{subfigure}{.2\columnwidth}
  \centering
  \includegraphics[width=\linewidth]{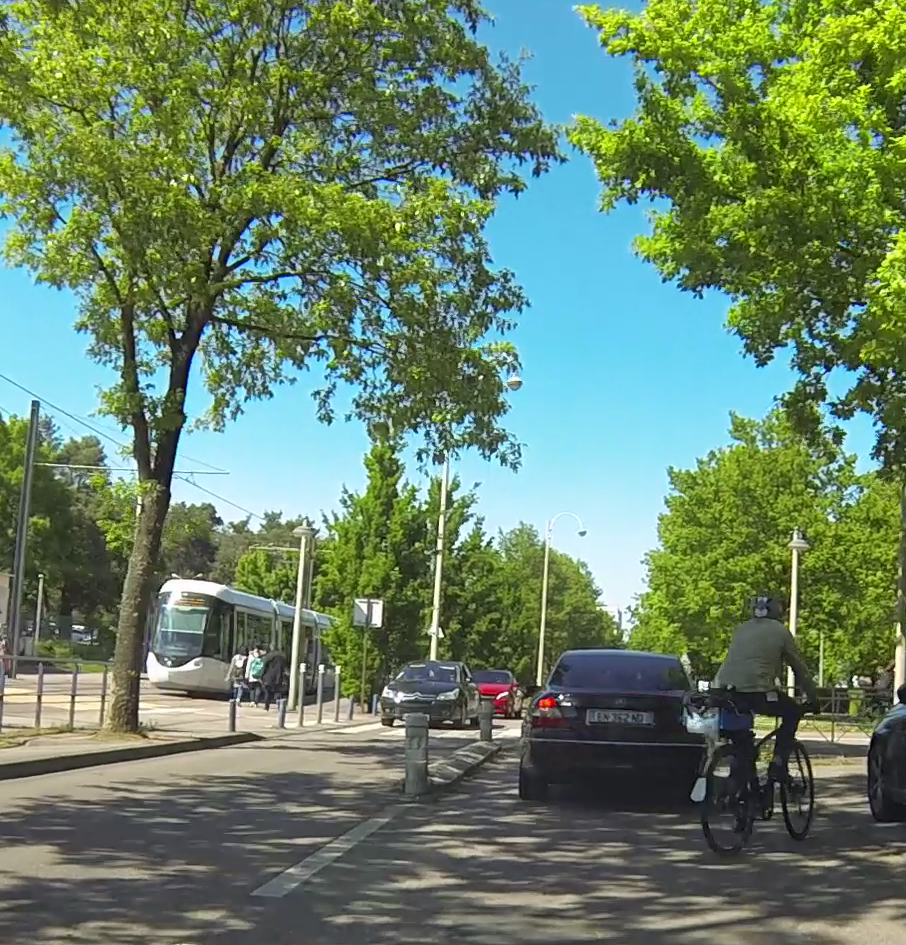}
\end{subfigure}
\caption{\rebuttalCor{}{Examples of images used to train the adapted CycleGAN. First row are the $I_0$ channels of some polarimetric images and second row are some RGB ones.}}
\label{fig:polar_example}
\end{figure}

To conduct the experiments, we rely on a combination of two datasets presented in \cite{blin2020new, blin2019road}, composed of polarimetric and RGB images.
From these datasets we select 2485 unpaired images from each domain (RGB and polarimetry). Example instances are shown in Figure~\ref{fig:polar_example}. The polarimetric images are of dimension $500 \times 500 \times 4$. The latter dimension is due to the four intensities acquired by the camera, namely $I_0, I_{45}, I_{90}$ and $I_{135}$. The RGB images are of dimension $906 \times 945 \times 3$.

% \begin{figure}
% \centering
% \begin{subfigure}{.2\columnwidth}
%   \centering
%   \includegraphics[width=\linewidth]{images/0030144.png}
% \end{subfigure}%
% \begin{subfigure}{.2\columnwidth}
%   \centering
%   \includegraphics[width=\linewidth]{images/0038544.png}
% \end{subfigure}%
% \begin{subfigure}{.2\columnwidth}
%   \centering
%   \includegraphics[width=\linewidth]{images/0025879.png}
% \end{subfigure}%
% \begin{subfigure}{.2\columnwidth}
%   \centering
%   \includegraphics[width=\linewidth]{images/0032059.png}
% \end{subfigure}%
% \begin{subfigure}{.2\columnwidth}
%   \centering
%   \includegraphics[width=\linewidth]{images/0088999.png}
% \end{subfigure}
% \caption{Examples of RGB images used to train the adapted CycleGAN.}
% \label{fig:rgb_example}
% \end{figure}
 
Our constrained CycleGAN is trained for 400 epochs on randomly cropped patches of size $200\times 200$. As for the constraints, we found experimentally that setting the hyper-parameters $\mu = 1$ and $\nu = 1$ in \eqref{eqn:lfinal} provides the best performances. The hyper-parameter $\lambda$, controlling the reconstruction cost, is set to $\lambda = 10$. The learning rate is decreased linearly from $2 \times 10^{-4}$ to $2 \times 10^{-6}$ during the 400 training epochs.

To evaluate the effectiveness of our trained generative model, we consider KITTI and BDD100K (only using daytime images since polarimetry fails to characterize objects during nighttime) which often serve as testbed in applications related to road scene object detection. The constrained-output CycleGANs we train, are used to transfer RGB images from KITTI and BDD100K to the polarimetric domain. The resulting datasets are denoted respectively as Polar-KITTI and Polar-BDD100K. Since the CycleGAN architecture is fully convolutional, it has no requirement on the input image's size. Therefore, even though the model was trained on $200 \times 200$ patches, it scales straightforwardly to the images of size $1250 \times 375$ from KITTI and of size $1280 \times 720$ from BDD100K datasets. \rebuttalCor{}{We also randomly horizontally flip, as flipping the image does not alter the physical properties of a polarimetric image \cite{blanchonPolarimetricImageAugmentation2021}.}

To assess whether or not fulfilling the physical constraints is paramount, we investigate a variant of Polar-KITTI and Polar-BDD100K: we learn a standard unconstrained CycleGAN based on the same unpaired RGB/polarimetric images.

\subsection{Evaluation of the generated images} \label{subsec:eval_gen_img}

\begin{figure*}[!th]
     \centering
    \begin{subfigure}{.11\linewidth}
  \centering
  \includegraphics[width=\linewidth]{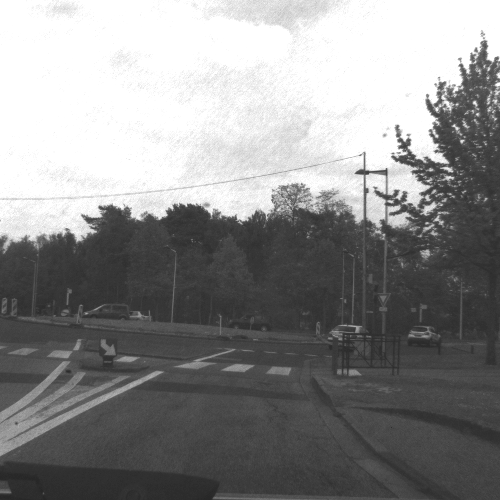}
\end{subfigure}%
\begin{subfigure}{.11\linewidth}
  \centering
  \includegraphics[width=\linewidth]{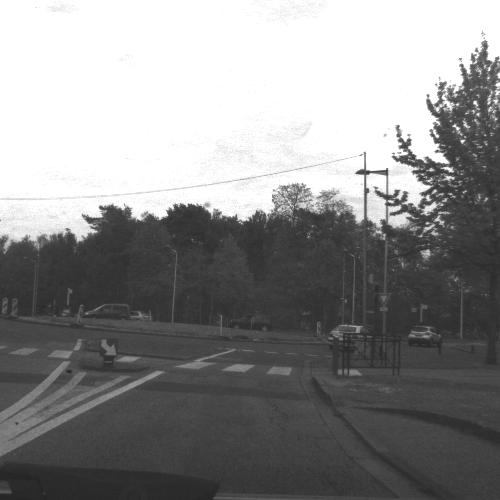}
\end{subfigure}%
\begin{subfigure}{.11\linewidth}
  \centering
  \includegraphics[width=\linewidth]{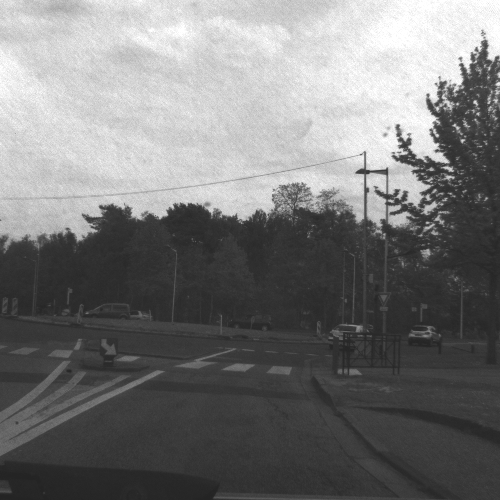}
\end{subfigure}%
\begin{subfigure}{.11\linewidth}
  \centering
  \includegraphics[width=\linewidth]{images/0001611_I0.png}
\end{subfigure}%
\begin{subfigure}{.105\linewidth}
  \centering
  \includegraphics[width=\linewidth]{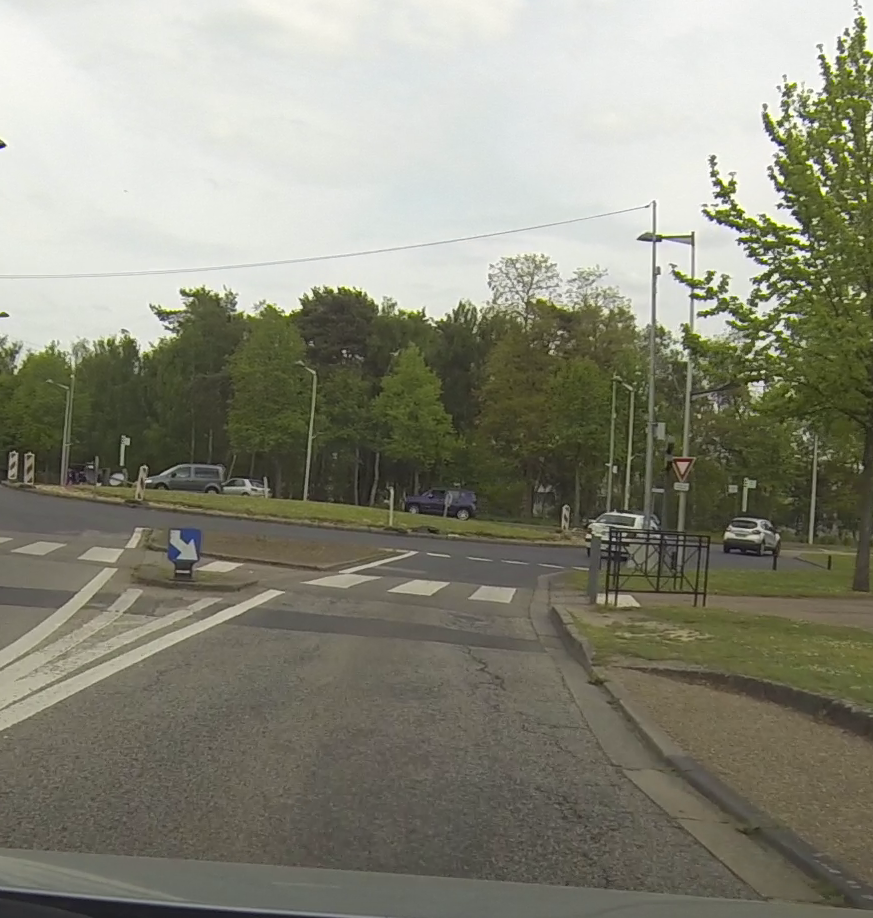}
\end{subfigure}%
\begin{subfigure}{.105\linewidth}
  \centering
  \includegraphics[width=\linewidth]{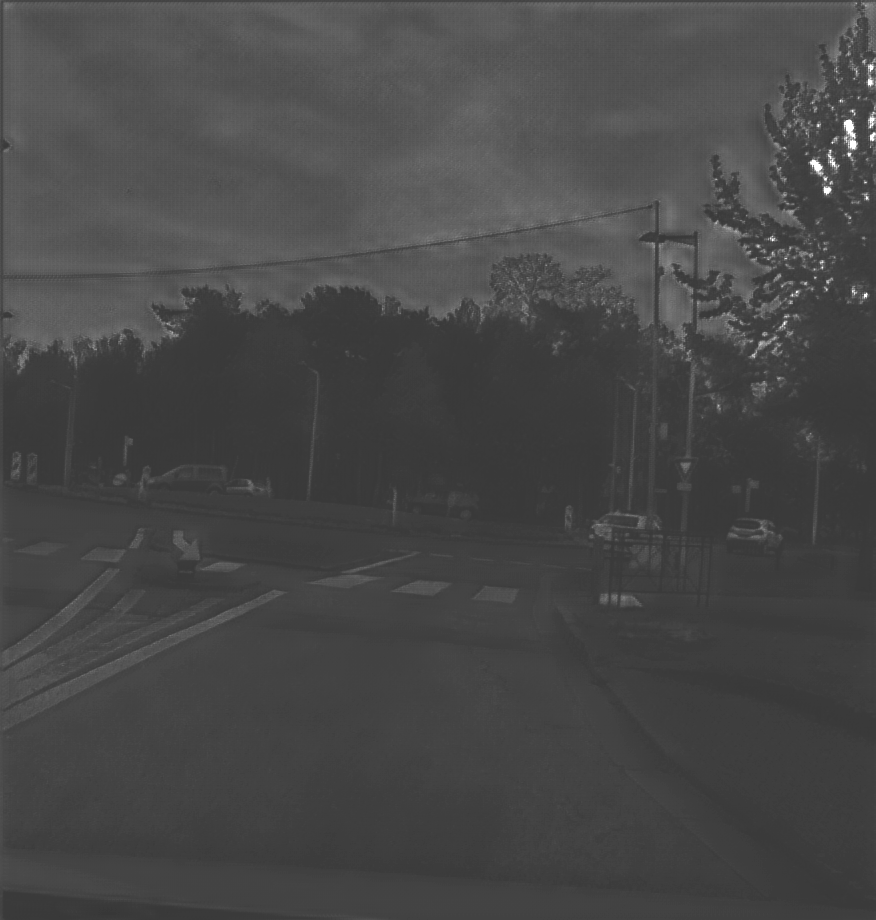}
\end{subfigure}%
\begin{subfigure}{.105\linewidth}
  \centering
  \includegraphics[width=\linewidth]{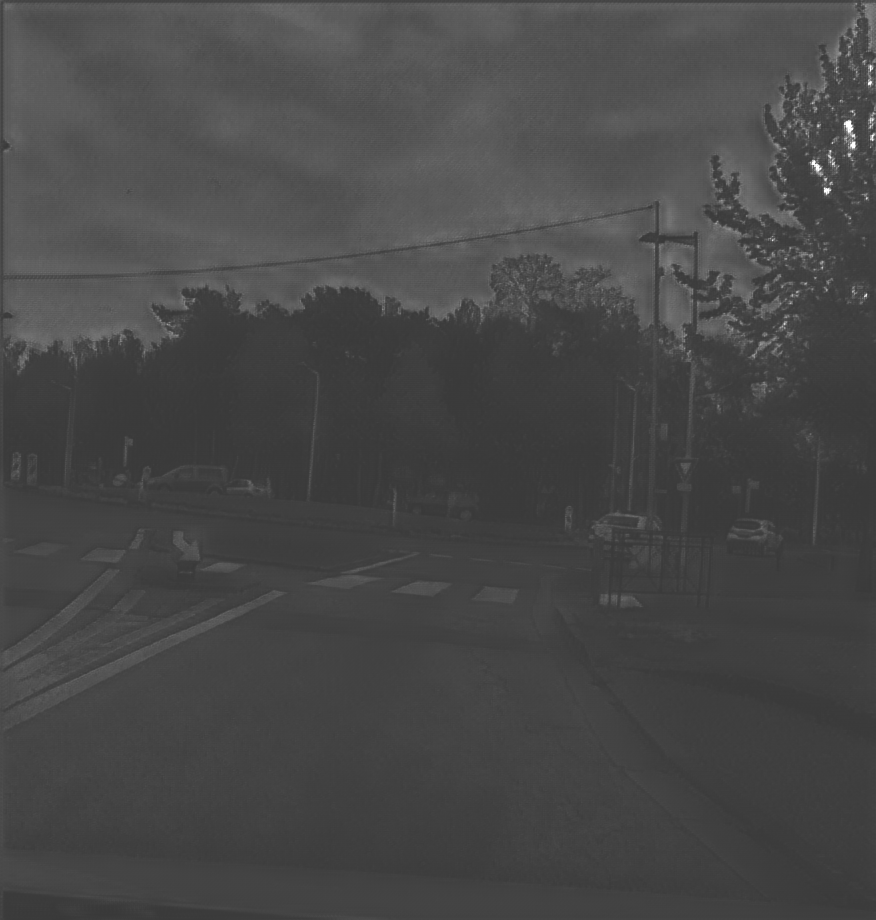}
\end{subfigure}%
\begin{subfigure}{.105\linewidth}
  \centering
  \includegraphics[width=\linewidth]{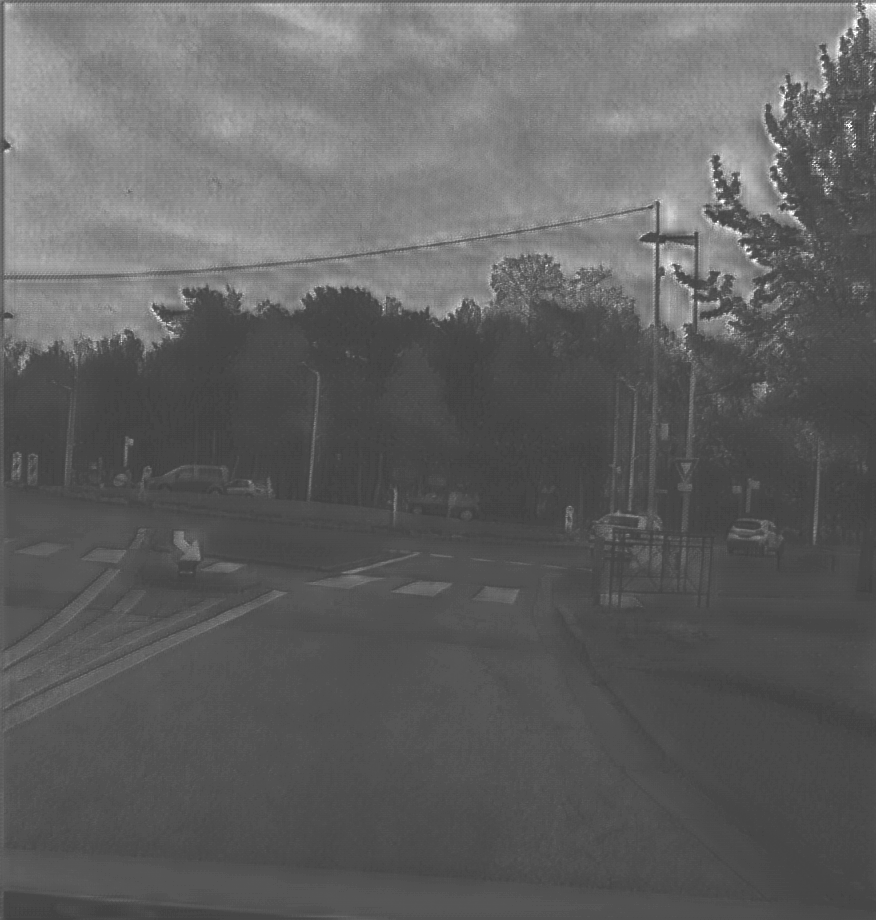}
\end{subfigure}%
\begin{subfigure}{.105\linewidth}
  \centering
  \includegraphics[width=\linewidth]{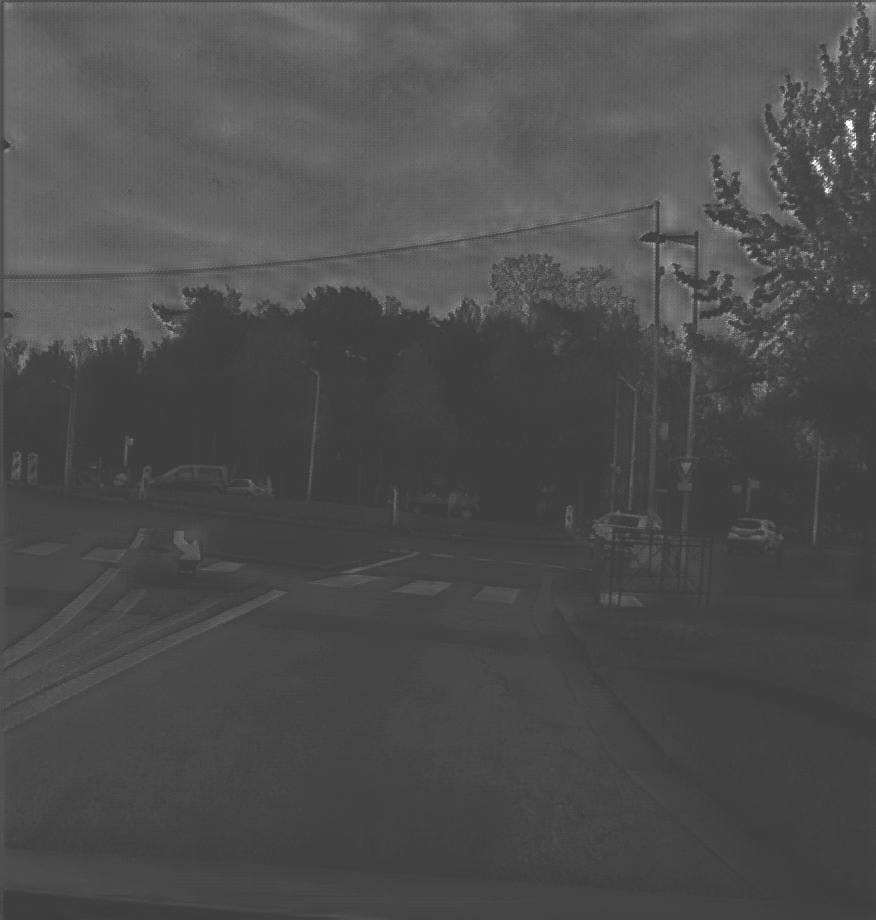}
\end{subfigure}
    \begin{subfigure}{.11\linewidth}
  \centering
  \includegraphics[width=\linewidth]{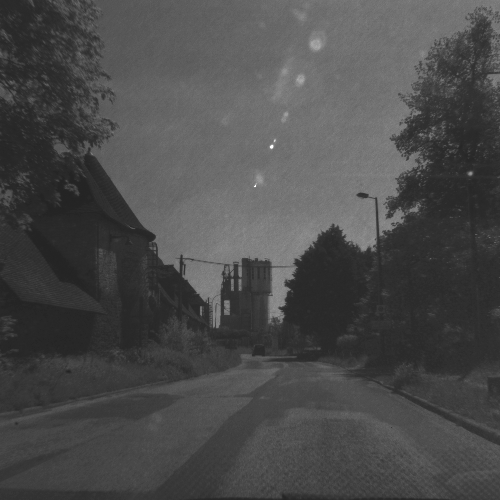}
\end{subfigure}%
\begin{subfigure}{.11\linewidth}
  \centering
  \includegraphics[width=\linewidth]{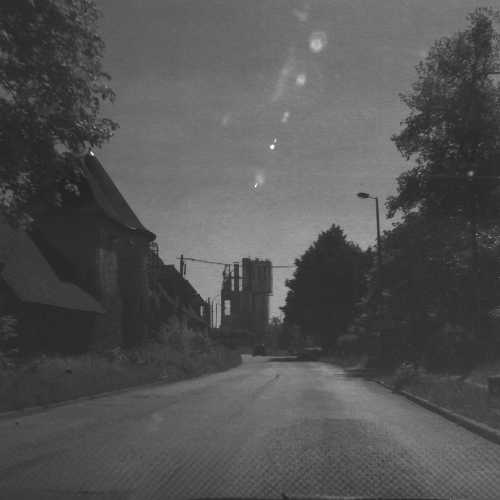}
\end{subfigure}%
\begin{subfigure}{.11\linewidth}
  \centering
  \includegraphics[width=\linewidth]{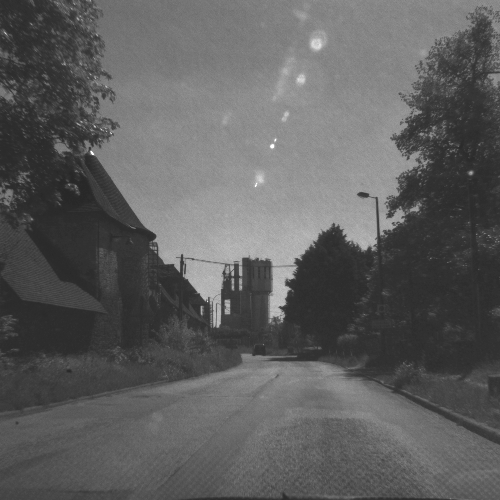}
\end{subfigure}%
\begin{subfigure}{.11\linewidth}
  \centering
  \includegraphics[width=\linewidth]{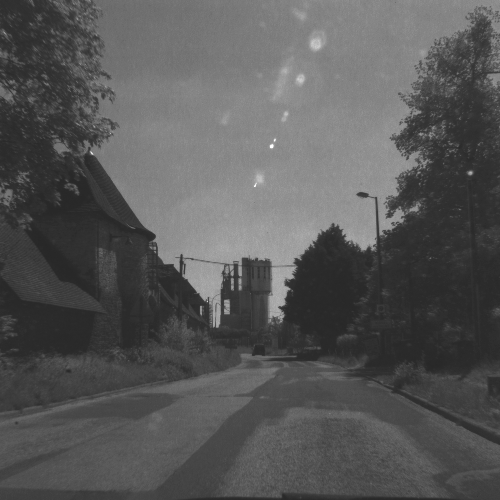}
\end{subfigure}%
\begin{subfigure}{.105\linewidth}
  \centering
  \includegraphics[width=\linewidth]{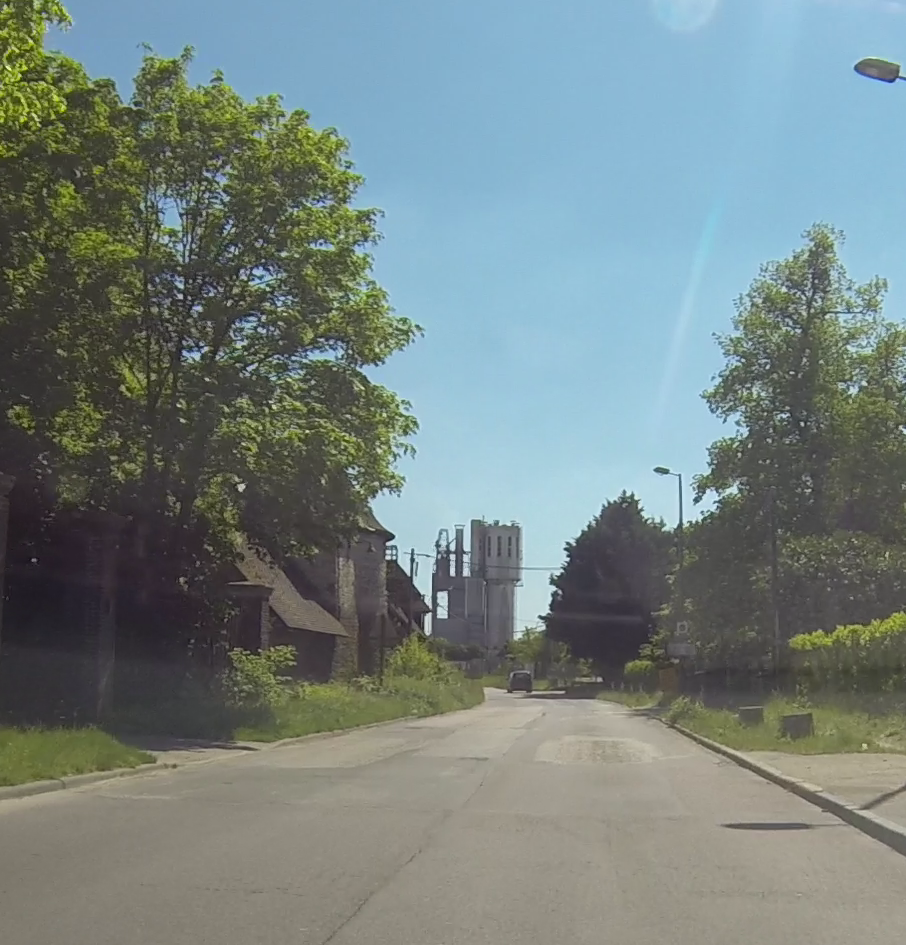}
\end{subfigure}%
\begin{subfigure}{.105\linewidth}
  \centering
  \includegraphics[width=\linewidth]{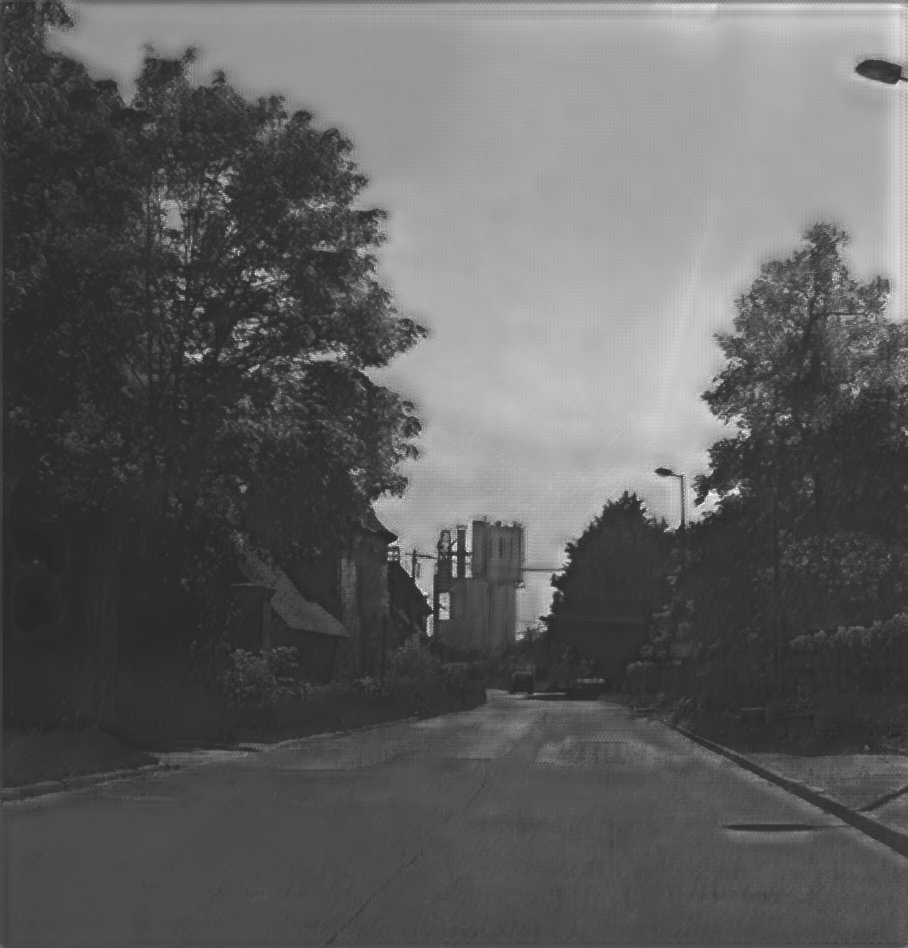}
\end{subfigure}%
\begin{subfigure}{.105\linewidth}
  \centering
  \includegraphics[width=\linewidth]{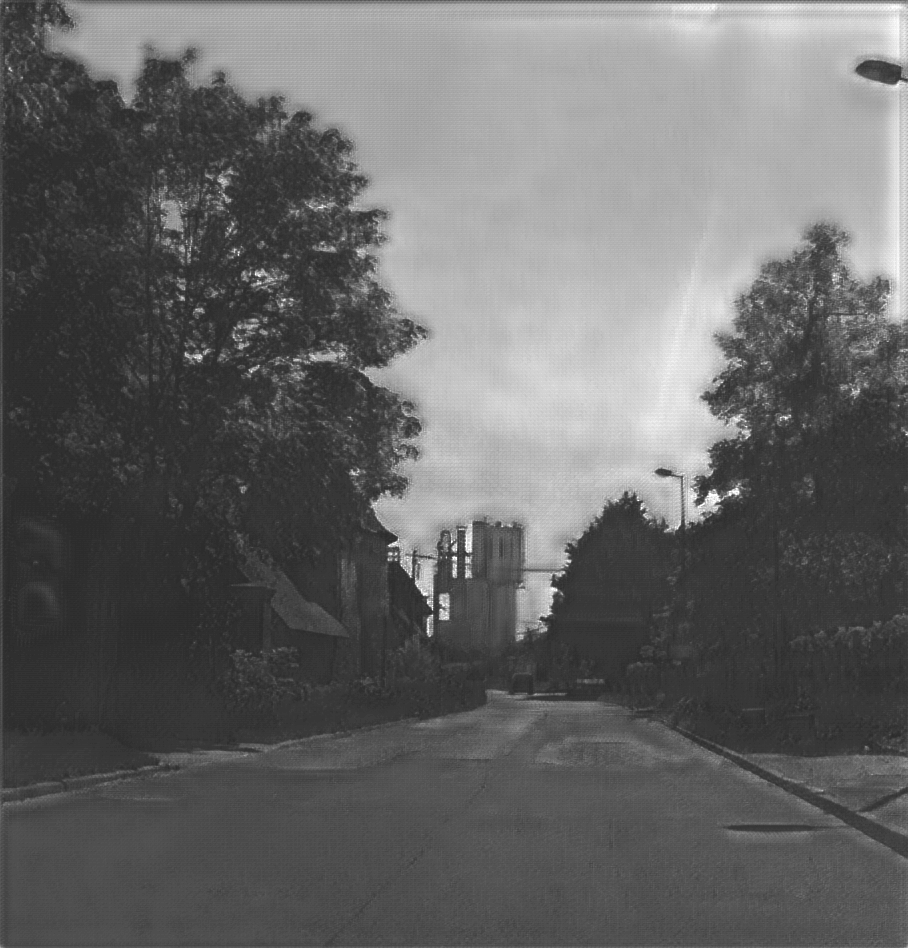}
\end{subfigure}%
\begin{subfigure}{.105\linewidth}
  \centering
  \includegraphics[width=\linewidth]{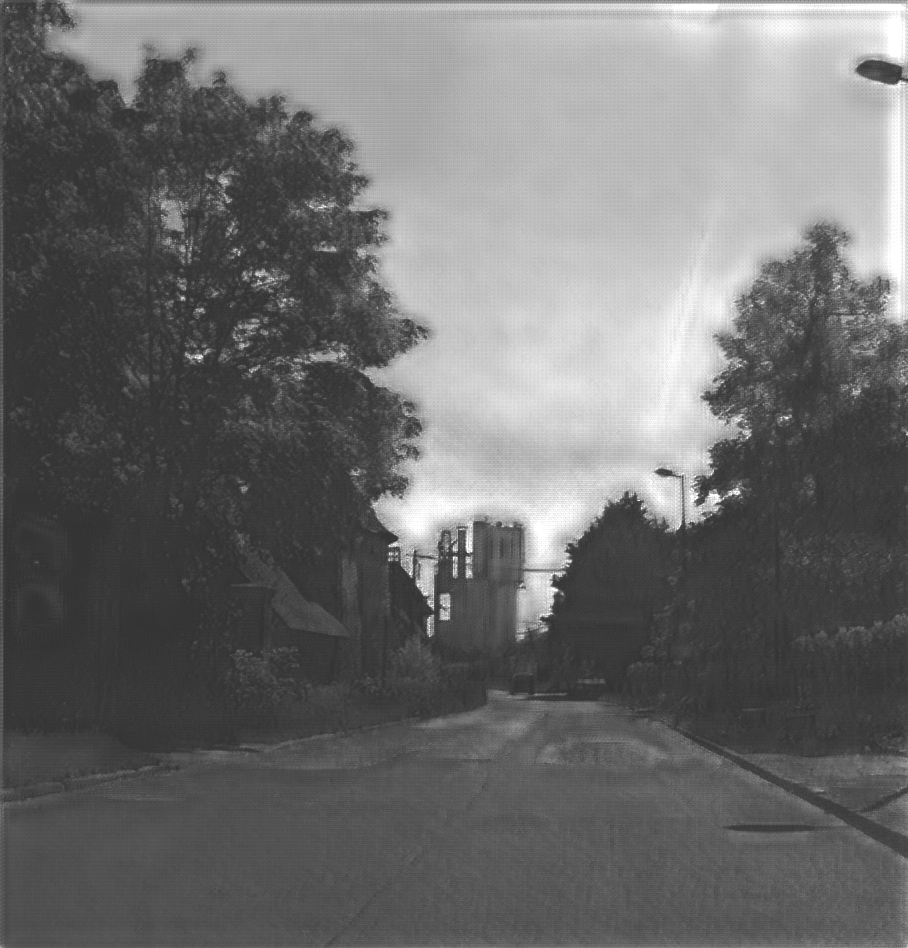}
\end{subfigure}%
\begin{subfigure}{.105\linewidth}
  \centering
  \includegraphics[width=\linewidth]{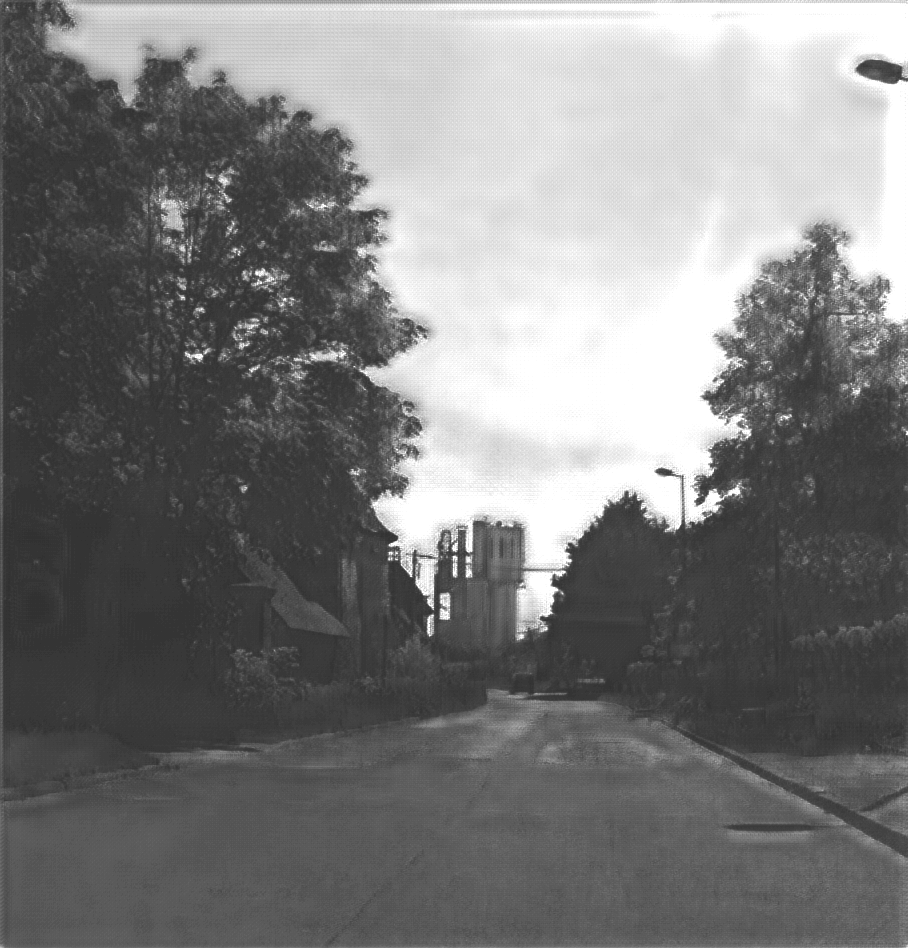}
\end{subfigure}
    \begin{subfigure}{.11\linewidth}
  \centering
  \includegraphics[width=\linewidth]{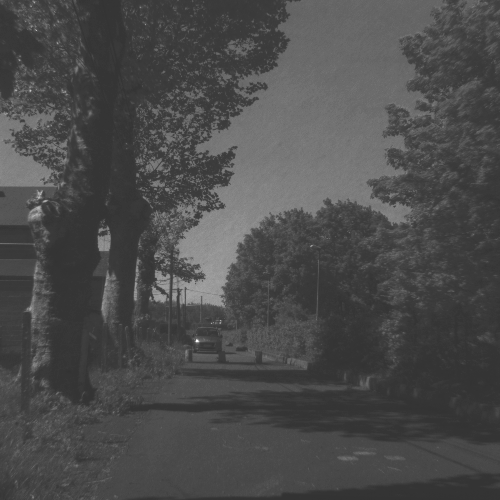}
\end{subfigure}%
\begin{subfigure}{.11\linewidth}
  \centering
  \includegraphics[width=\linewidth]{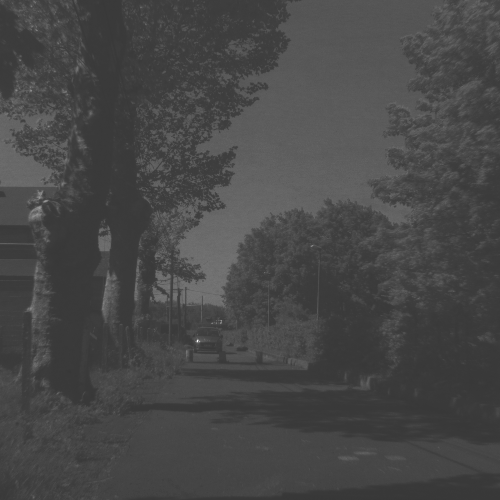}
\end{subfigure}%
\begin{subfigure}{.11\linewidth}
  \centering
  \includegraphics[width=\linewidth]{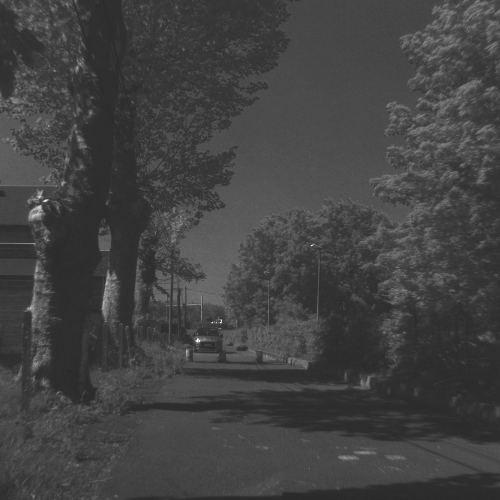}
\end{subfigure}%
\begin{subfigure}{.11\linewidth}
  \centering
  \includegraphics[width=\linewidth]{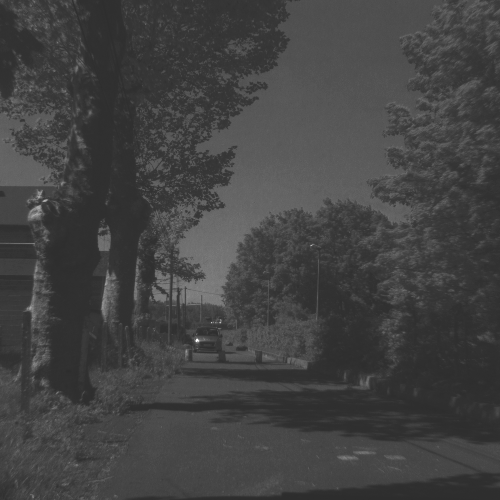}
\end{subfigure}%
\begin{subfigure}{.105\linewidth}
  \centering
  \includegraphics[width=\linewidth]{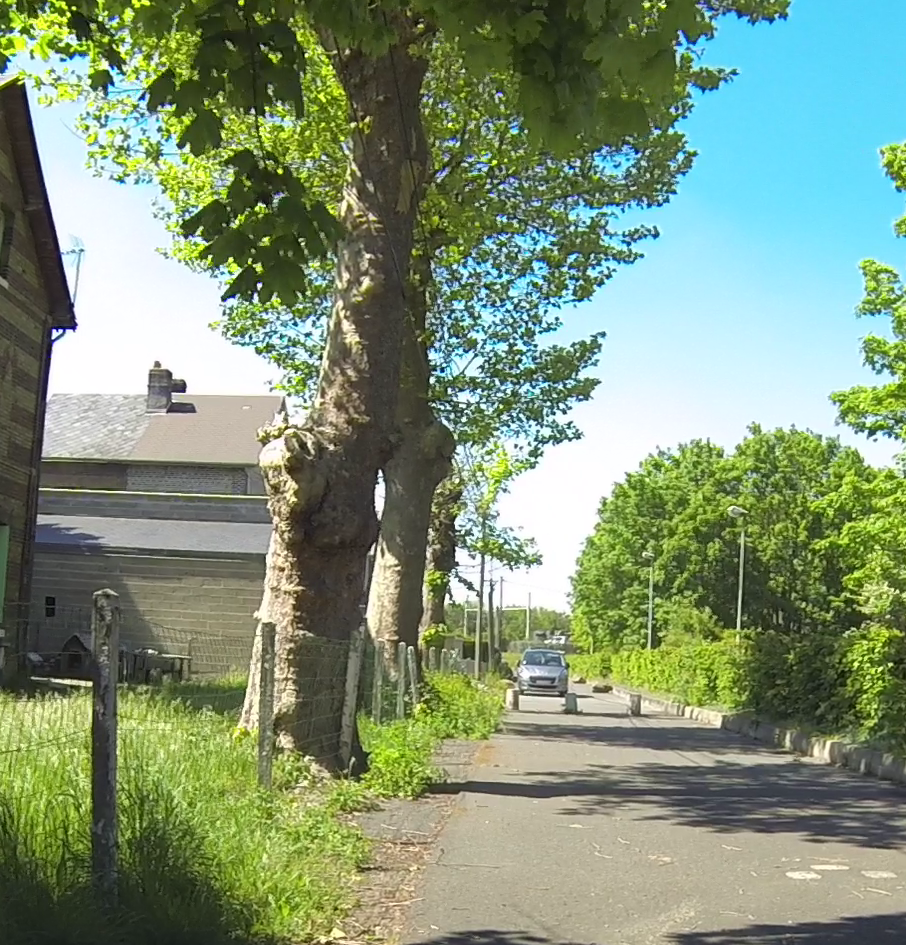}
\end{subfigure}%
\begin{subfigure}{.105\linewidth}
  \centering
  \includegraphics[width=\linewidth]{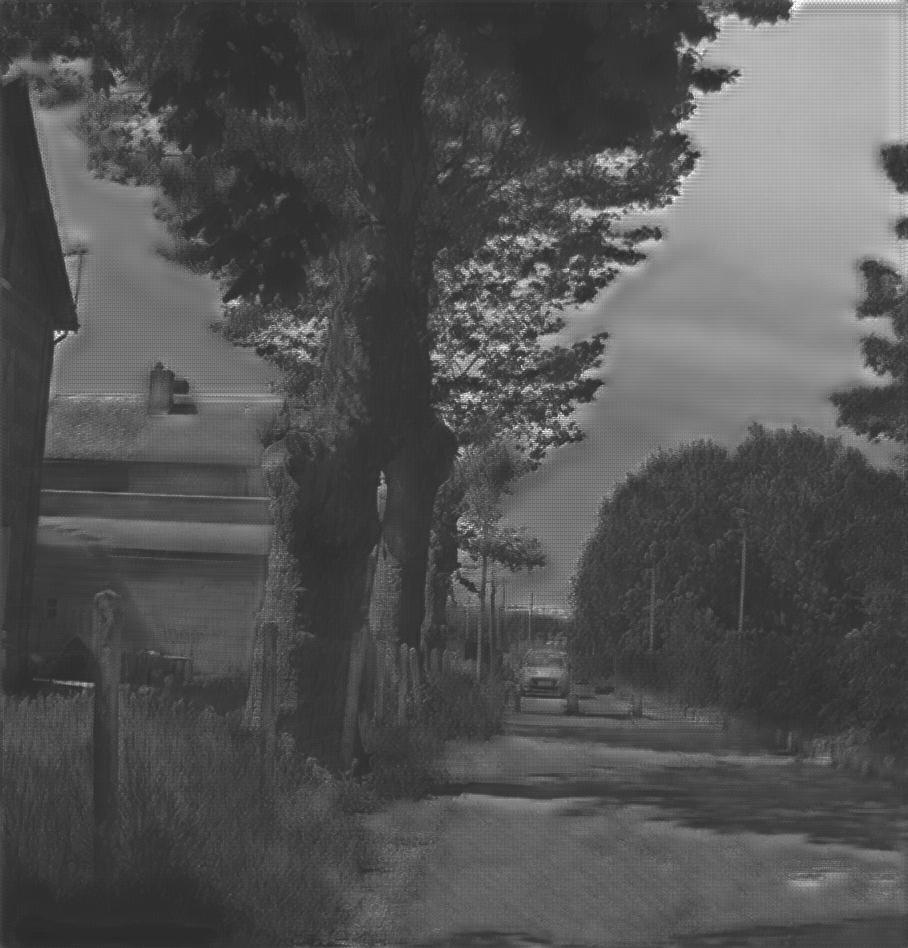}
\end{subfigure}%
\begin{subfigure}{.105\linewidth}
  \centering
  \includegraphics[width=\linewidth]{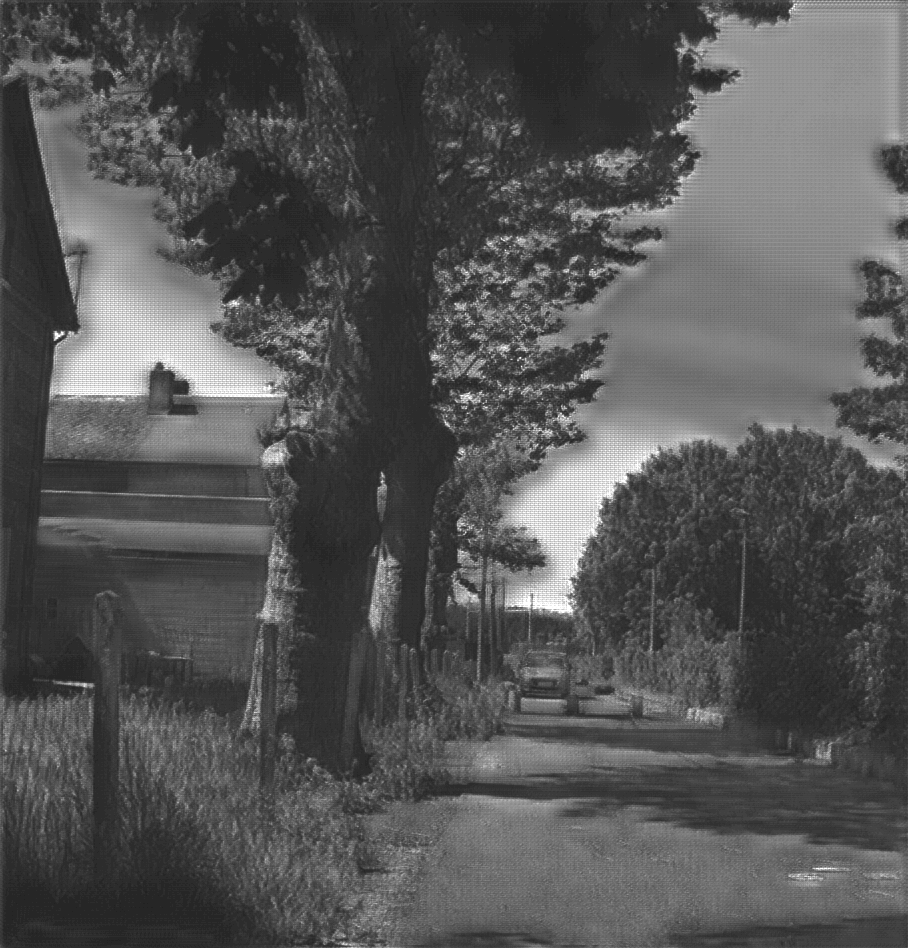}
\end{subfigure}%
\begin{subfigure}{.105\linewidth}
  \centering
  \includegraphics[width=\linewidth]{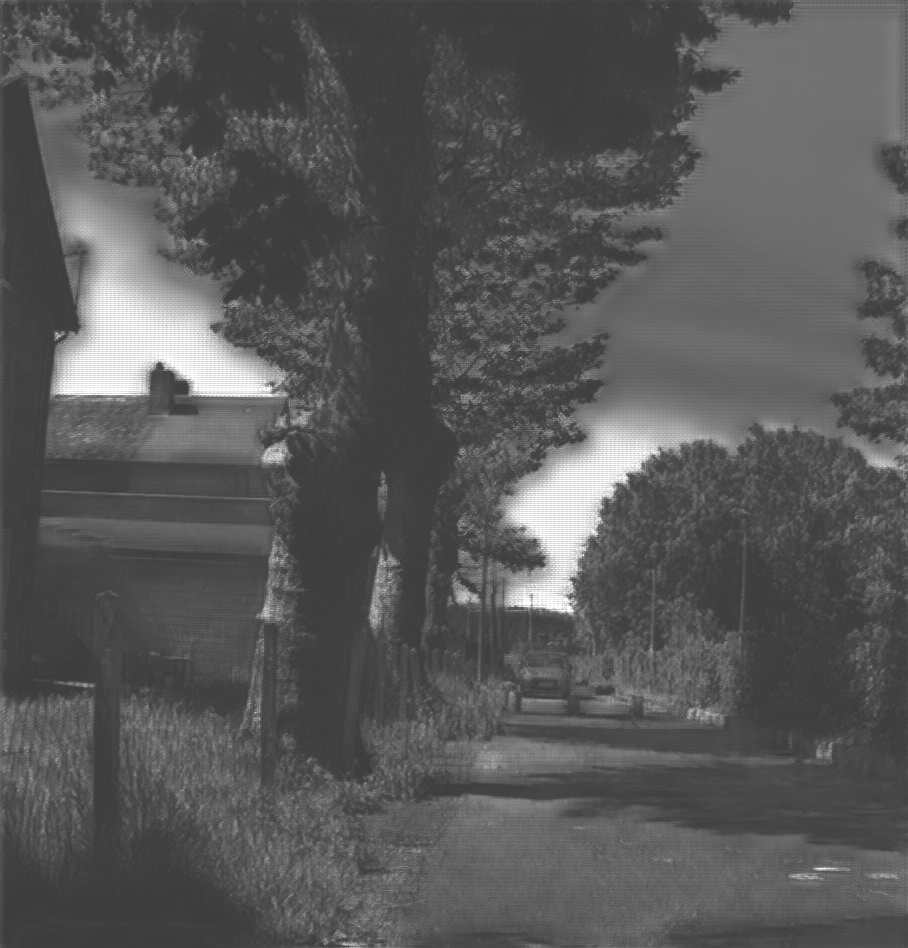}
\end{subfigure}%
\begin{subfigure}{.105\linewidth}
  \centering
  \includegraphics[width=\linewidth]{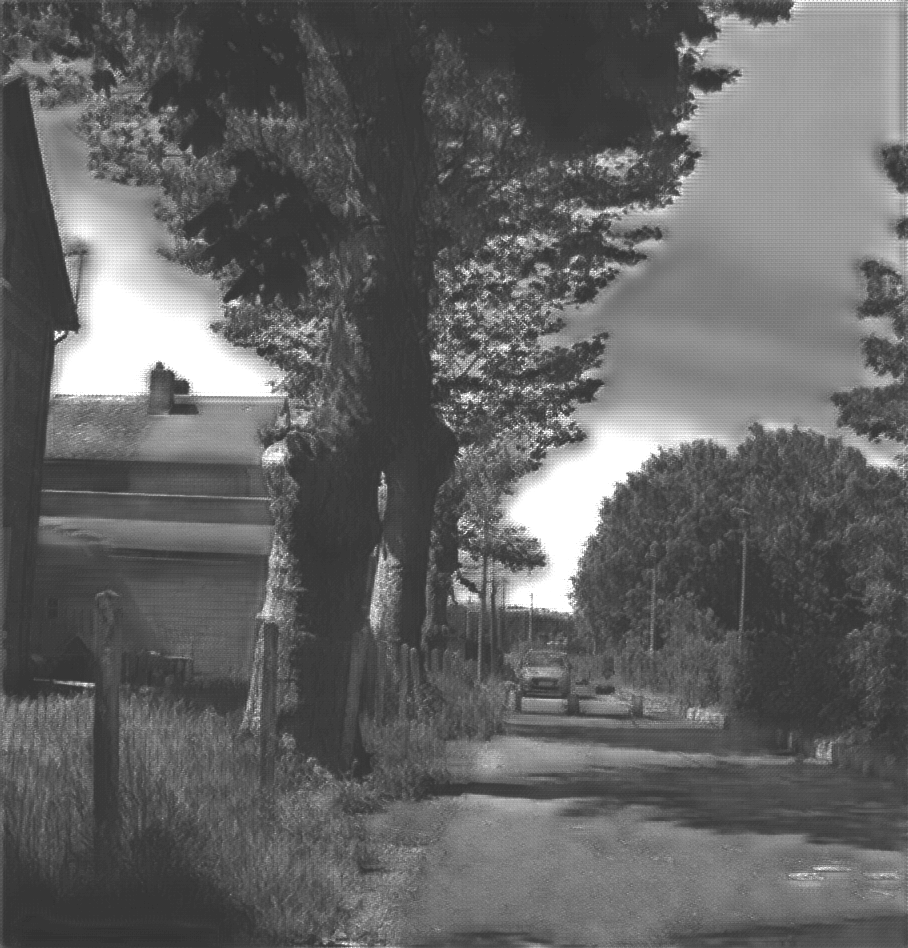}
\end{subfigure}
    \begin{subfigure}{.11\linewidth}
  \centering
  \includegraphics[width=\linewidth]{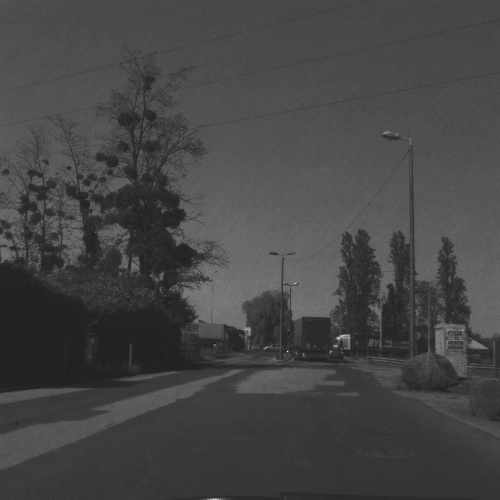}
\end{subfigure}%
\begin{subfigure}{.11\linewidth}
  \centering
  \includegraphics[width=\linewidth]{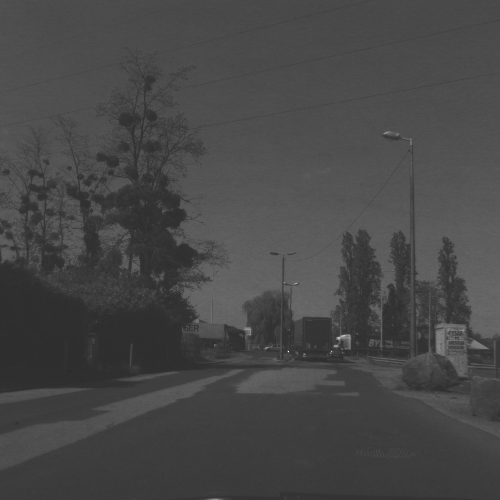}
\end{subfigure}%
\begin{subfigure}{.11\linewidth}
  \centering
  \includegraphics[width=\linewidth]{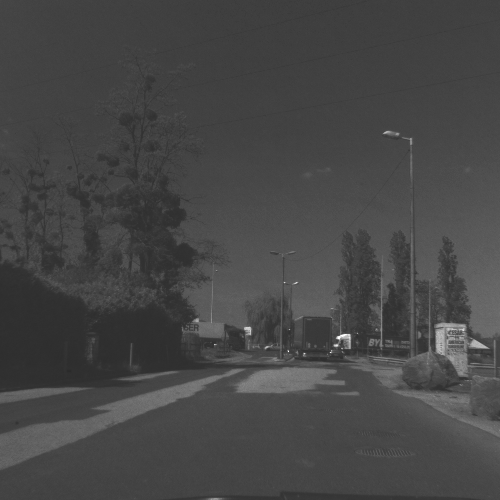}
\end{subfigure}%
\begin{subfigure}{.11\linewidth}
  \centering
  \includegraphics[width=\linewidth]{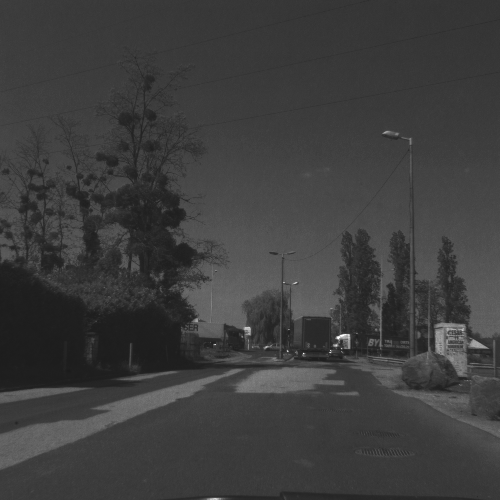}
\end{subfigure}%
\begin{subfigure}{.105\linewidth}
  \centering
  \includegraphics[width=\linewidth]{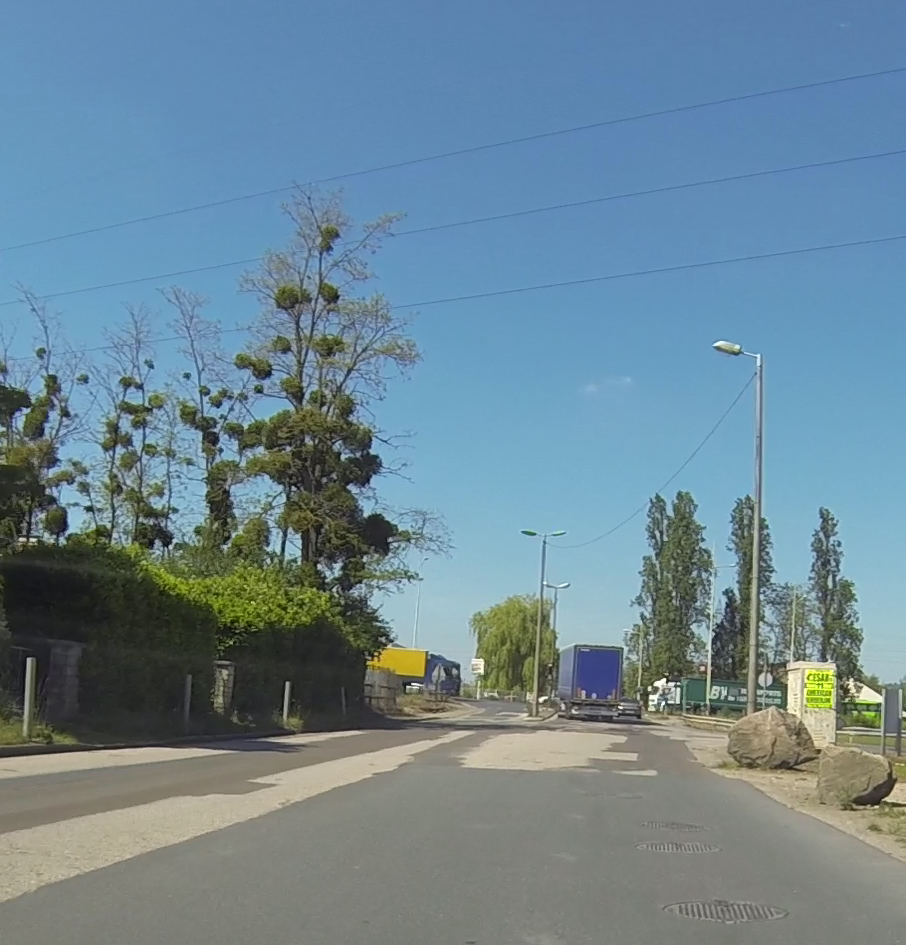}
\end{subfigure}%
\begin{subfigure}{.105\linewidth}
  \centering
  \includegraphics[width=\linewidth]{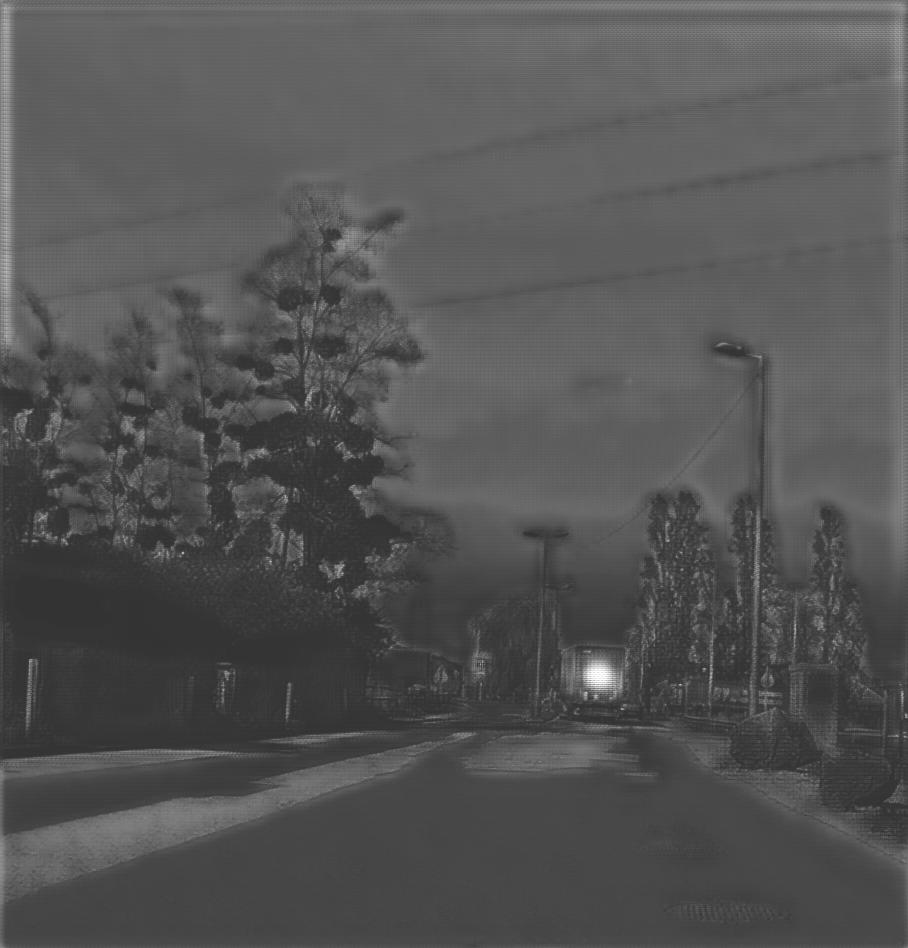}
\end{subfigure}%
\begin{subfigure}{.105\linewidth}
  \centering
  \includegraphics[width=\linewidth]{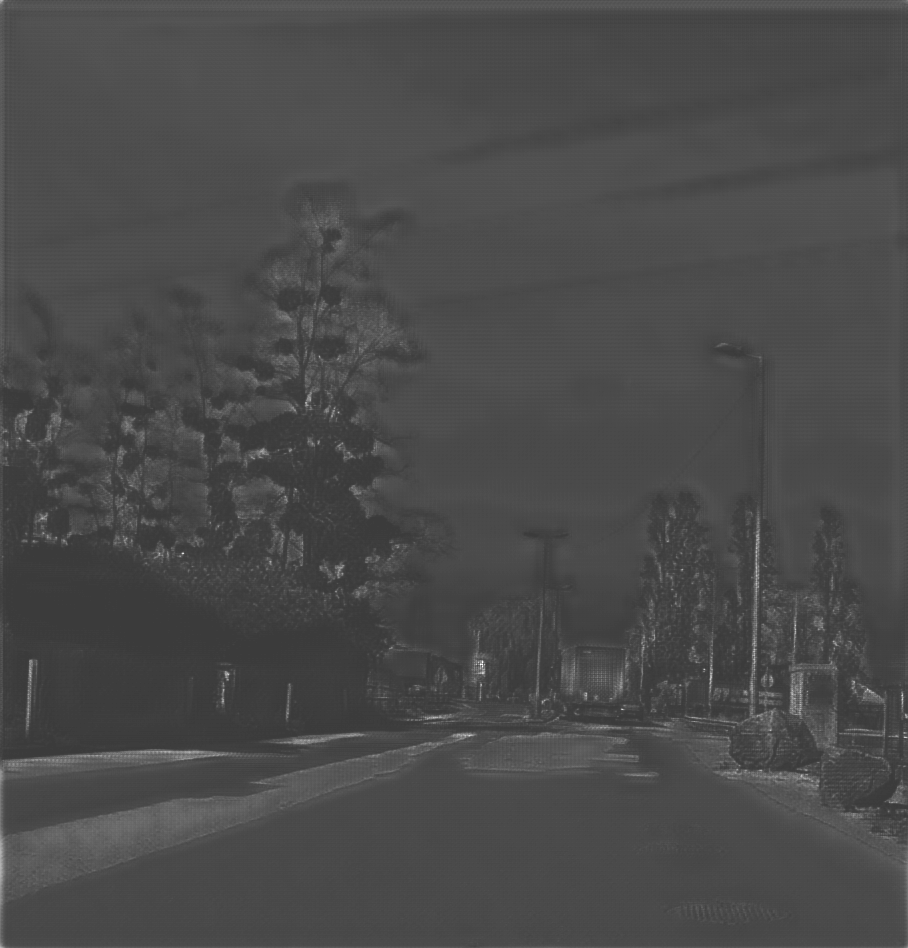}
\end{subfigure}%
\begin{subfigure}{.105\linewidth}
  \centering
  \includegraphics[width=\linewidth]{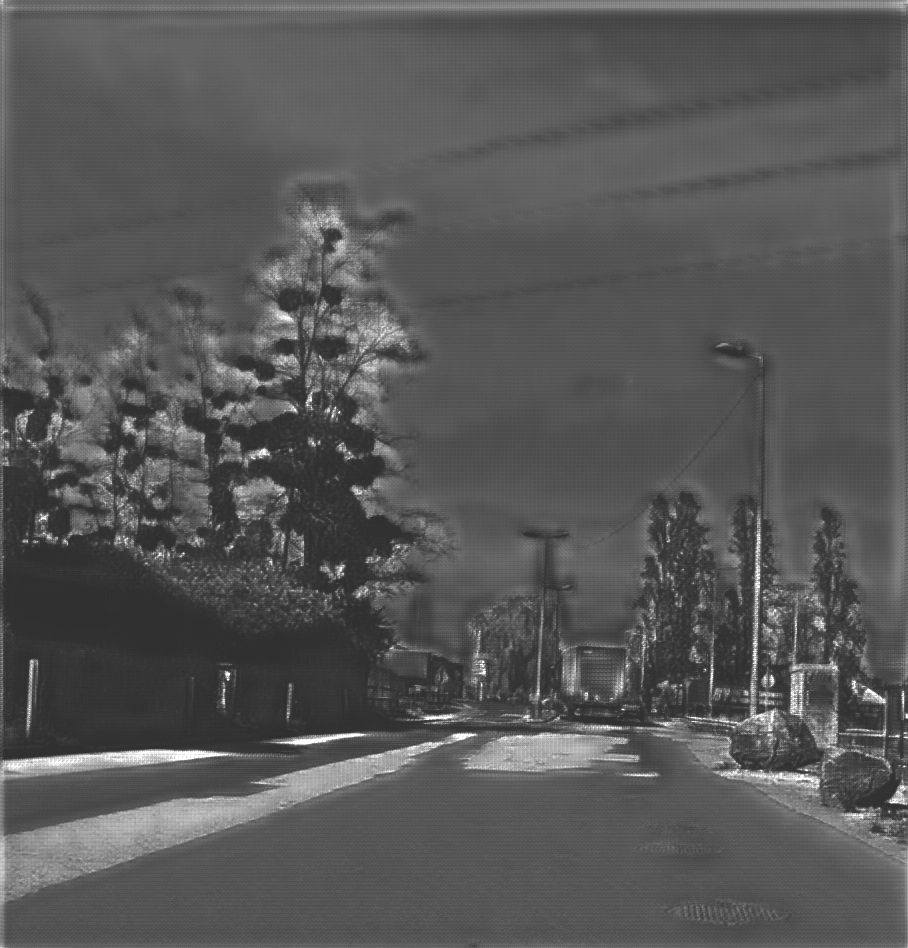}
\end{subfigure}%
\begin{subfigure}{.105\linewidth}
  \centering
  \includegraphics[width=\linewidth]{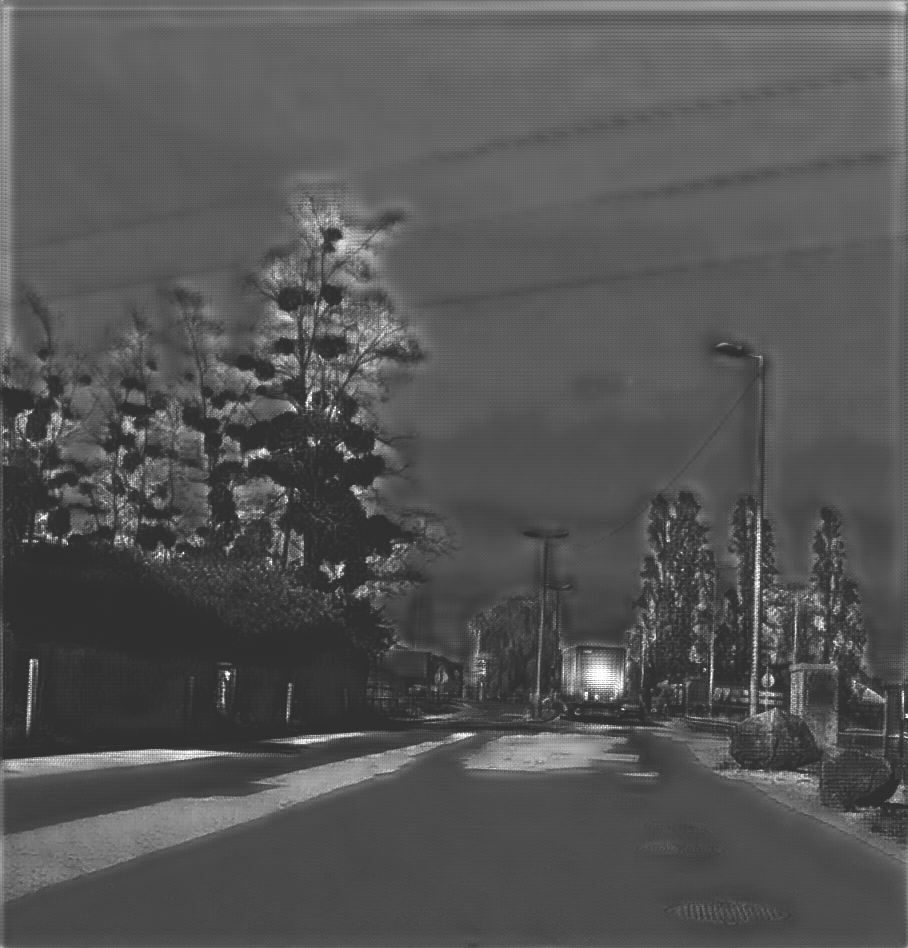}
\end{subfigure}
    \caption{Examples of generated polarimetric images. From left to right: $I_0$, $I_{45}$, $I_{90}$, $I_{135}$ (ground truth), RGB image and $I_0$, $I_{45}$, $I_{90}$, $I_{135}$ generated from RGB.}
    \label{fig:reco_polar}
\end{figure*}

In order to assert the ability of the generated Polar-KITTI and Polar-BDD100K datasets to preserve the relevant features for road scene applications, we train a detection network following the setup in Figure~\ref{fig:experimental_setup}. For this experiment, a RetinaNet-50 \cite{lin2017focal} pre-trained on the MS COCO dataset \cite{lin2014microsoft} is fine-tuned in three different settings. In the first setup, the detection model is fine-tuned on the original RGB KITTI (or BDD100K) while the second one considers fine-tuning on the generated polarimetric images from KITTI (Polar-KITTI) or BDD100K (Polar-BDD100K) datasets. The third one uses the unconstrained variant of the generated images from KITTI or BDD100K datasets. After the first fine-tuning, the three final detection models are fine-tuned one last time on the real polarimetric dataset (described in Table \ref{tab:dataset_properties}). 
\begin{table}[!th]
\begin{center}
\begin{tabular}{c c c c c}
    \specialrule{.2em}{.1em}{.1em}
    Class & Train & Val & Test \\
    \specialrule{.2em}{.1em}{.1em}
    Images & 3861 & 1248 & 509 \\
    \specialrule{.2em}{.1em}{.1em}
    car & 19587 & 3793 & 2793 \\
    person & 2049 & 294 & 161 \\
    bike & 16 & 35 & 3 \\
    motorbike & 52 & 4 & 5 \\
\end{tabular}
\caption{Polarimetric dataset features. From the third row, the total number of instances within each class are indicated.}
\label{tab:dataset_properties}
\end{center}
\end{table}
Overall, the trained CycleGAN and detection networks under these settings are evaluated in qualitative and quantitative ways. The end goal is to check: (i) the ability of the generated images to help learning polarimetry-based features for object detection, and (ii) the influence of respecting the polarimetric feasibility constraints on detection performances.

We measure the visual quality of the generated images by computing the classical Fréchet Inception Distance \cite{heusel2017}. Computing this distance requires to extract visual features from each set of images (real and generated) using a pre-trained deep neural network (usually an Inception v3 \cite{szegedy2016} network pre-trained on ImageNet \cite{deng2009imagenet}) and to evaluate the Fréchet (or Wasserstein) distance between the distributions of these features, which are assumed to be Gaussian distributions. We evaluate this distance using 509 images from each generated polarimetric dataset and from the test set as described in Table \ref{tab:dataset_properties}.

As for the feature extractor, since the classical Inception v3 network is not adapted to polarimetric images, we use the convolutional part of a polarimetry-adapted RetinaNet detection network \cite{blin2019road}, which has been trained on the MS-COCO dataset and fine-tuned on the same real polarimetric dataset that we used in our CycleGAN experiments.
In order to evaluate the improvements in the detection, we compute the error rate evolution $ER_o$. The improvement $ER_o$ on the detection of the object $o$ is given by:
\begin{equation}
    \label{eq:error_rate}
    ER_o = \frac{(1 - AP_o^{p}) - (1 - AP_o^{RGB})}{1-AP_o^{RGB}}\enspace,
\end{equation}

\noindent where $AP_o^{RGB}$ and $AP_o^{p}$ respectively denote the average precision for object $o$ detection in RGB and in polarimetric images. A negative $ER_0$ means that $AP_o^{p}$ was improved over $AP_o^{RGB}$.

\subsection{Results and discussion}

\begin{figure*}[!th]
    \centering
    \includegraphics[width=0.8\linewidth]{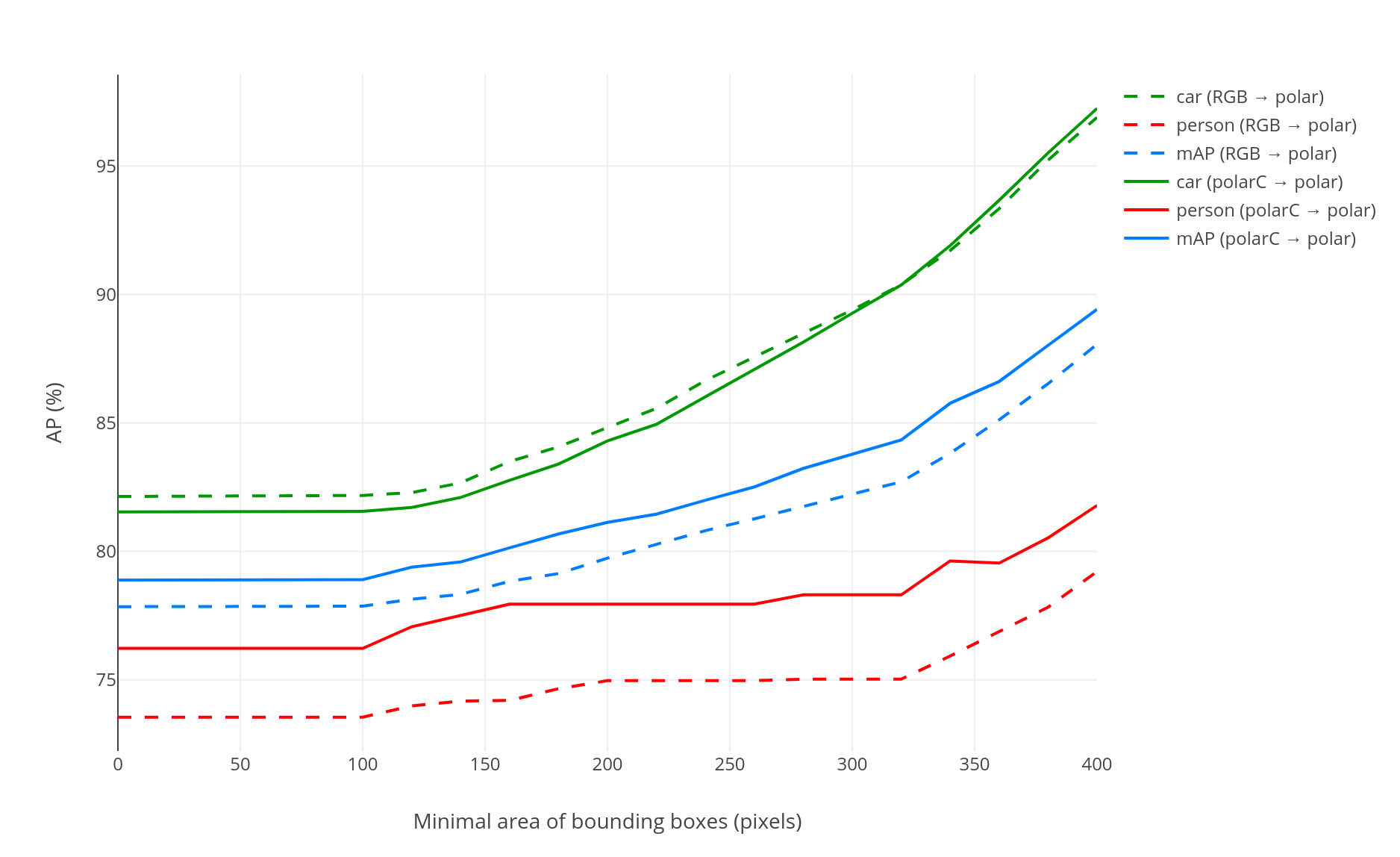}
    \caption{Evolution of the average precision (AP) when setting a minimal area of the detected bounding boxes. Here green, blue and red lines respectively refer to the evolution of cars', $mAP$ and person's detections. The dashed and the solid lines respectively refer to the use of BDD100K RGB and Polar-BDD100K during training.}
    \label{fig:bounding_boxes}
\end{figure*}

First the qualitative coherence of the generated images is evaluated. The polarimetric images generated from their RGB equivalent are shown in Figure~\ref{fig:reco_polar}. 

\rebuttalAdd{

\begin{table*}[!th]

\begin{center}

\begin{tabular}{c c c c| c c cc}
    \specialrule{.3em}{.2em}{.2em}
    Dataset & Class & Test & $ER_o$ & Dataset & Class & Test & $ER_o$ \\
    \specialrule{.3em}{.2em}{.2em}
    KITTI RGB & person & 0.663 & N/A & BDD100K RGB & person & 0.736 & N/A \\
    & car & 0.785 & N/A & & car & \textbf{0.821} & N/A \\
    & $mAP$ & 0.724 & N/A & & $mAP$  & 0.778 & N/A \\
    \specialrule{.2em}{.1em}{.1em} 
    unconstrained & person & 0.673 & -0.03 &  unconstrained  & person & 0.720 & 0.06 \\
    Polar-KITTI & car & 0.786 & -0.01 & Polar-BDD100K & car & 0.816 & 0.03 \\
    & $mAP$ & 0.730 & -0.02 & & $mAP$  & 0.768 & 0.05 \\
    \specialrule{.2em}{.1em}{.1em} 
    Polar-KITTI & person & \textbf{0.704} & -0.12 & Polar-BDD100K & person & \textbf{0.762} & -0.10 \\
     with contraints & car & \textbf{0.794} & -0.04 & with constraints & car & 0.815 & 0.03 \\
    & $mAP$  & \textbf{0.749} & -0.09 & & $mAP$  & \textbf{0.789} & -0.05 \\
\end{tabular}
\caption{
\rebuttalCor{}{Comparison of the detection performances for the \textit{person} and \textit{car} classes, as well as the mean average precision (mAP). RetinaNet-50 pre-trained on MS COCO is the baseline of all the experiments. The first row is the fine-tunning on the original RGB KITTI/BDD100K. The second row is the fine-tuning on Polar-KITTI/Polar-BDD100K without constraints and the bottom row is the detection models fine-tuned on Polar-KITTI/Polar-BDD100K with the constraints. For a given dataset (KITTI or BDD100K), the best results overall  are shown in boldface}
}
\label{tab:obtained_results}
\end{center}
\end{table*}
\begin{table}[]
\begin{center}
\begin{tabular}{c c c c}
    \specialrule{.3em}{.2em}{.2em}
    Datasets & $\mathcal{C}$ & Mean & Median \\
    \specialrule{.3em}{.2em}{.2em}
    Real & $\mathcal{C}_1$ & 0.06 $\pm$ 0.04 & 0.04 \\
    polar & $\mathcal{C}_2$ & 2.47 $\pm$ 7.11\% & 0.48\% \\
    \specialrule{.2em}{.1em}{.1em} 
    unconstrained & $\mathcal{C}_1$ & 0.26 $\pm$ 0.19 & 0.23 \\
    Polar-KITTI & $\mathcal{C}_2$ & 27.31 $\pm$ 43.5\% & 2.15\% \\
    \specialrule{.2em}{.1em}{.1em} 
    Polar-KITTI & $\mathcal{C}_1$ & 0.12 $\pm$ 0.04 & 0.12 \\
     with constraints & $\mathcal{C}_2$ & 1.55 $\pm$ 3.36\% & 0.14\% \\
\end{tabular}

\end{center}
\caption{\label{tab:polar_constraints}
Evaluation of the constraint fulfillment using the designed losses $L_{\mathcal{C}_1}$ and $L_{\mathcal{C}_2}$ at the pixel scale. \rebuttalCor{}{Here, we compare the real polarimetric dataset to the generated polarimetric KITTI datasets, constrained and unconstrained. The column $\mathcal{C}$ indicates the evaluated constraint, with $\mathcal{C}_1$ referring to the calibration constraint ($I = AS$) and $\mathcal{C}_2$   to the optical constraint ($S_0^2 \geqslant S_1^2 + S_2^2$).} The mean and the median of the percentage of pixels in an image that do not fulfill the constraints $\mathcal{C}_2$ is computed. Regarding the constraint $\mathcal{C}_1$, we normalize it by computing the mean and the median of $||I - AS|| / (||I|| + ||AS||)$.}
\end{table}
}
As for the constraints, Table \ref{tab:polar_constraints} shows how including them to the CycleGAN's loss helps generating images which better fulfill the physical polarimetric properties at the pixel scale. The errors related to the constraints $\mathcal{C}_1$ and $\mathcal{C}_2$ on generated images are consistent with the observed measurement errors on the real images, whereas the unconstrained approach yields poor results. Constraint $\mathcal{C}_3$ is met for all generated images thanks to the use of hyperbolic tangent as activation function for the last layer of the generative models. Additionally, the obtained Fréchet Inception Distances are of \textbf{6022.7} for the unconstrained CycleGAN and \textbf{4485.1} for our approach. Note that the scale of the FID scores computed with the pre-trained RetinaNet is larger than when using a pre-trained Inception v3 network. Thus, these scores are to be interpreted as relative metrics. This indicates that taking the constraints into account improves visual and physical quality of the generated samples.

Next, as an example, we show the benefits of the generated images through an object detection task. This enables to check if objects contained in the generated scene are %globally
physically coherent. 
A RetinaNet-based detection model is learned according to the setups described in Section \ref{subsec:eval_gen_img} and the obtained detection performances in term of mean average precision ($mAP$) are summarized in Table \ref{tab:obtained_results}. We chose not to evaluate the bike and motorbike detection performances as the polarimetric dataset does not contain enough objects of those two classes.

As we can see in Table \ref{tab:obtained_results}, using the generated polarimetric images improves the detection performance in real polarimetric images. The improvement is substantial for car and pedestrian detection. We achieve a \rebuttalCor{}{4\% improvement of the error rate (see Equation \ref{eq:error_rate})} for car detection and of 12\% for pedestrian detection which leads to a global improvement of 9\% in the detection, using Polar-KITTI with constraints. Similarly, for the Polar-BDD100K dataset, we notice an improvement of 10\% for pedestrian detection which leads to an increased $mAP$ of 5\% (pedestrians and cars). However, we notice that, for BDD100K, similar detection performances are obtained either for RGB or polarimetric images. This is due to the fact that generated images using CycleGAN tends to lack precision on small objects. In order to asses the impact of this effect, we compare the evolution of the detection scores when varying the minimal area of the bounding boxes from which objects are taken into account for the detection task. The obtained results are shown for the Polar-BDD100K and the RGB BDD100K datasets are shown in Figure~\ref{fig:bounding_boxes} and illustrate that, when the minimal area of bounding boxes increases, the $AP$ of car regarding the training including Polar-BDD100K, overcomes the one including RGB BDD100K. Even if the results are limited by the lower quality of the small objects in the generated images in this specific case, we can conclude that the generated polarimetric images help improving the overall detection results.

\section{Conclusion and future work}

In this work, we propose an efficient way to generate realistic polarimetric images subject to physical admissibility constraints. We design two loss terms that help enforcing these constraints and propose to adapt the CycleGAN algorithm to achieve the generation of physically admissible images. To train the proposed output-constrained CycleGAN, we combine the standard CycleGAN's objective function with two designed cost functions in order to handle the feasibility constraints related to each polarization-encoded pixel in the image. 
With the proposed generative model, we successfully translate RGB images from road scenes to polarimetric images showing an enhancement of the detection performances.
Future work would consist in improving the quality of the small objects in generated images. Another extension could be the generation of polarimetric images to other domains such as medical and Synthetic-Aperture Radar (SAR) imaging. Additionally, to strictly ensure the physical feasibility constraints, solutions such as directly addressing the optimization task involved by the genuine constrained CycleGAN problem, instead of its proposed relaxation, using methods such as proximal gradient descent method could be explored. \rebuttalAdd{Finally, since the constraints we develop are not limited to the generative methods, they might be applicable to some other augmentation techniques, such as rotation or noise.}

\section{Acknowledgements}
This  work  is  supported  by  the  ICUB  project  2017  ANRprogram  :  ANR-17-CE22-0011.  We  also  thank our  colleagues at  CRIANN, who provided us with the computation resources necessary for our experiments.

\printbibliography
\vspace{0.5cm}
% \section*{Supplementary Material}

% Supplementary material that may be helpful in the review process should
% be prepared and provided as a separate electronic file. That file can
% then be transformed into PDF format and submitted along with the
% manuscript and graphic files to the appropriate editorial office.

\end{document}